\documentclass[11pt]{article}
\usepackage[utf8]{inputenc}
\usepackage[T1]{fontenc}
\usepackage[margin=1in]{geometry}
\usepackage{lmodern}
\usepackage{amsmath,amssymb}
\usepackage{graphicx}
\usepackage{booktabs,longtable,array,calc}
\usepackage[numbers,sort&compress]{natbib}
\usepackage{xcolor}
\usepackage[colorlinks=true,linkcolor=blue!50!black,citecolor=blue!50!black,urlcolor=blue!50!black]{hyperref}
\usepackage{fvextra}
\usepackage[section]{placeins}  

\fvset{fontsize=\small}
\RecustomVerbatimEnvironment{verbatim}{Verbatim}{breaklines=true,fontsize=\small,frame=single}
\title{Learning through Discernment: when a small language model learns from user feedback on the job, and when it decays}
\author{Sirisilp Kongsilp\\[2pt]
\small Decentralized Science, Inc.\\
\small Department of Computer Engineering, Faculty of Engineering, Kasetsart University\\
\small \texttt{tai@kratai.com}, \texttt{sirisilp.k@ku.th}}
\date{November 2026}

\begin{document}
\maketitle
\begin{abstract}
Small language models on devices could keep improving by learning, overnight, from their own daily work. Without a true outcome, however, something must decide which attempts are worth learning from, and both the model's own judgment and the user's reaction can be wrong. We study a small model (Qwen3 0.6B and 1.7B) with a LoRA adapter, fine-tuned each night on the attempts a judge accepts, over 127 lives of up to a year on WebShop and $\tau^2$-bench. We compare the true outcome, the user's stated verdict and the model's own judgment with a separate, frozen ``feedback sense'' that reads the user's reaction, built in three versions. Judging itself, the model improved and then decayed, over a year and whatever its adapter. A sense that took polite give-ups for success collapsed 3 of 5 lives; one built for discernment, which first asks whether the user gave a clear verdict, kept 4 of 5 near the oracle for two weeks, then faded on clean but scarce lessons, to the level of self-judging by day 60. When the user cannot see the outcome ($\tau^2$-bench), the sense read reactions correctly, yet 80\% of its lessons were failures the customer had thanked. Across 97 lives, the share of failed attempts among the lessons predicted the outcome (Spearman $\rho$ = -0.81). In our experiments, learning from users held up only where the judge kept failures out and the user could see the outcome; lopsided lessons may also narrow the model.
\end{abstract}

\section{Introduction}

Small language models now run on phones, laptops and local servers, and once released they meet tasks their training did not anticipate. Retraining them centrally is slow, and it would mean sending the user's interactions away from the device. A model that improved from its own daily work, learning overnight and on the device, would close that gap: small models built for devices \cite{liu2024mobilellm} and fine-tuning on the device itself \cite{zhu2023pockengine, zheng2025review} make this setting realistic. It is also where learning on the job matters most, and it is the setting we study, with models of 0.6B and 1.7B parameters (Section 4).

The hard part is not the training but the judgment. Learning on the job needs something to decide which of the day's attempts were good enough to learn from. Without a true outcome there are two obvious candidates: the model's own judgment of its work \cite{yuan2024selfrewarding}, or the user's reaction to it \cite{hancock2019learning, shi2026wildfeedback}. Both can be wrong. Models judge their own work poorly \cite{huang2024large}, and users are often polite about failures, thanking the assistant for trying and moving on. A wrong lesson is not merely wasted: once trained on, it teaches the model to repeat the failure.

Most work on self-training and on learning from feedback measures one or a few rounds of improvement on a fixed benchmark \cite{zelikman2022star, singh2024beyond, yuan2024selfrewarding}. Little measures what happens over weeks of nightly updates, where each night's model learns from what the previous nights taught it, and where a judge's small errors can compound.

We study that setting with a ``creature'': a frozen small model with a LoRA adapter that works by day and is fine-tuned each night on the attempts a judge accepts, replaying its recent lessons. We compare judges that stand in for published approaches, under one loop, the same tasks and the same training budget: the true outcome (an \emph{oracle} and upper bound), the user's stated verdict read without error, and the model's own reflection on its attempt, with and without the user's words. As an alternative to the model judging itself, we build a separate, frozen ``feedback sense'' that reads the user's reaction to each attempt. We built it in three versions, each meant to fix a weakness of the one before. What it must provide we call \emph{discernment}: telling an informed, clear verdict on the result from a polite, vague or uninformed reaction. We ran five experiments on WebShop \cite{yao2022webshop} and $\tau^2$-bench \cite{barres2026tau2}, with lives of 14 days to a year: 127 lives in all.

Learning from its own judgment, the creature improved and then decayed, over 14 days and over a year, and whatever the design of its adapter. The first sense held up with a user who reacts plainly to the outcome. The second, facing users who react as people do, took polite give-ups for success, and three of its five lives collapsed after learning from them on the first night. The third, which first asks whether the user gave a clear verdict at all, kept four of its five lives near the \emph{oracle} for two weeks, then faded slowly although its lessons stayed clean, to the level of self-judging by day 60. On $\tau^2$-bench, where the customer cannot see whether its task was done, the sense read the customers' reactions correctly, yet 80\% of what it learned were failures that the customers had thanked the agent for. Across the 14-day WebShop lives, one number predicted where a life ended: the share of failed attempts among its lessons ($\rho$ = -0.81 across 97 lives).

Our contributions are:
\begin{enumerate}
\item
  A long-horizon test of nightly self-training for small models: five experiments, 127 lives of up to a year, two benchmarks and an \emph{oracle} upper bound, with all data and code public.
\item
  Evidence that the precision of a life's lessons goes with how it ends, and over two weeks tracks the outcome more closely than their number does: a few accepted failures can collapse a life. Precision is not enough, however: over longer lives, scarce lessons also faded.
\item
  A feedback sense with discernment, built in three versions, and the failure mode it addresses: polite reactions to failure.
\item
  Two limits of learning from users: a reaction carries no information the user did not have, and even correct lessons can fade when they are scarce, and may narrow the model's skills when they are lopsided.
\end{enumerate}

Section 2 places the work among related research; Section 3 describes the creature and the feedback sense; Section 4 the experimental setup; Section 5 the results, one experiment per subsection; and Sections 6-8 the discussion, limitations and conclusion.

\section{Related work}

Our setting sits where three lines of work meet: continual learning, learning from human feedback, and self-training. It also draws on work on learning after deployment, on language-model agents and on small models that run on a device.

\textbf{Continual learning and forgetting.} Neural networks trained on a sequence of tasks tend to lose earlier skills, a problem known since early connectionist work \cite{mccloskey1989catastrophic}. Remedies include penalising changes to the weights that mattered for earlier tasks \cite{kirkpatrick2017overcoming} and replaying stored experience \cite{rolnick2019experience}; De Lange et al.\ survey the field \cite{delange2022continual}. Language models forget during continual fine-tuning too \cite{luo2025empirical}, and parameter-efficient adapters such as LoRA \cite{hu2022lora} learn less but also forget less than full fine-tuning \cite{biderman2024lora}. This work almost always assumes a labelled stream of tasks. Our creature has no labels: it learns from its own daily work, and a judge decides what enters the stream. We keep a LoRA with replay of recent lessons (Section 3) and test adapters built to protect old learning (Experiment 1-C); the forgetting we see in Experiment 5 is of the kind this literature describes, caused here by lessons that cover the skills unevenly.

\textbf{Learning from human feedback.} Language models are commonly aligned with human preferences collected offline, through a learned reward model \cite{christiano2017deep, ouyang2022training}, with feedback from another model \cite{bai2022constitutional}, or directly from pairs of preferred and rejected outputs \cite{rafailov2023dpo}. KTO \cite{ethayarajh2024model} needs only a binary signal per example, whether an output was desirable, which is the closest in form to our yes or no lessons. In all of these the labels are explicit and gathered before training. We study the opposite setting: the signal is the user's implicit reaction, gathered while the model works, and it has to be interpreted before anything can be learned from it.

\textbf{Self-training and self-judgment.} Training on a model's own verified successes is a long-standing recipe: expert iteration \cite{anthony2017thinking}, STaR \cite{zelikman2022star}, rejection-sampling fine-tuning \cite{yuan2023scaling}, ReST \cite{gulcehre2023reinforced} and ReST-EM \cite{singh2024beyond} generate attempts, keep those a checker accepts, and train on them. Our \emph{oracle} judge is a simplified member of this family. When no checker exists, the model can judge itself: self-rewarding models \cite{yuan2024selfrewarding} and their meta-judging extension \cite{wu2025meta} use the model's own evaluation as the reward. There are known reasons to doubt such judges. Models struggle to correct their own reasoning without outside feedback \cite{huang2024large}, language models used as judges show systematic biases \cite{zheng2023judging}, optimising a proxy reward eventually lowers the true one \cite{gao2023scaling}, and training on a model's own outputs can degrade it over generations \cite{shumailov2024ai}. Our \emph{self} and \emph{both} judges are simplified stand-ins for this family, and their decline over weeks (5.1) is consistent with these concerns.

\textbf{Learning after deployment from implicit feedback.} Closest to our work are systems that keep learning from their users after deployment. Hancock et al.\ \cite{hancock2019learning} let a chatbot extract new training examples from its own conversations, and later chatbots learned from feedback collected in deployment \cite{shuster2022blenderbot, xu2023learning}. In machine translation, implicit signals from users proved more useful than explicit star ratings \cite{kreutzer2018can}. Natural feedback is common in real chats and can be extracted at scale \cite{doneyehiya2024naturally, shi2026wildfeedback, peng2026wildreward}; the last two are among the datasets our sense learned from. Retrospective learning \cite{chen2025retrospective} reads the implicit signals in past interactions to improve a model over rounds, and MemoryBench \cite{ai2026memorybench} measures how well systems learn from simulated user feedback over time. These works show that implicit feedback can help. We ask a narrower question they leave open: over many nightly updates, who should decide which reactions become lessons, and what happens when the reactions cannot be trusted.

\textbf{Agents and their benchmarks.} Our creature is a tool-using agent in the style of ReAct \cite{yao2023react}. Reflexion \cite{shinn2023reflexion} lets an agent reflect on its failed attempts with outside feedback, which our \emph{both} judge mirrors. Agents have also been tuned on their own trajectories \cite{zeng2024agenttuning, xi2025agentgym}, including with DPO on pairs of failed and successful attempts on WebShop \cite{song2024trial}, close to our retry pairs. We use WebShop \cite{yao2022webshop} and $\tau^2$-bench \cite{barres2026tau2}, which extends $\tau$-bench \cite{yao2025taubench} to tasks where the customer acts too. Our WebShop user knows the outcome of each purchase, while $\tau^2$-bench's simulated customer sees only the conversation, not the database; Experiment 5 tests what that means for learning from reactions.

\textbf{Small and on-device models.} Small models designed for devices \cite{liu2024mobilellm, yang2025qwen3} and fine-tuning on the device itself \cite{zhu2023pockengine} make learning on the job possible where a user's data never leaves the device; Zheng et al.\ review language models at the edge \cite{zheng2025review}. This is why we study 0.6B and 1.7B models (Section 4).

\textbf{Sleep and consolidation.} The day and night framing echoes complementary learning systems \cite{mcclelland1995why}, in which fast learning during experience is consolidated offline. Two recent papers bring sleep to language models: one consolidates what was in context into the weights, with synthetic ``dreaming'' \cite{behrouz2026language}; the other uses offline computation between inputs to improve later inference \cite{lee2026do}. Both study what a sleep phase does with what the model has seen. We study the earlier question of who decides what it should learn from.

\textbf{Why no direct comparison with published results.} The published methods behind our judges were evaluated over one round of improvement on fixed benchmarks, not over weeks of nightly updates, and no long-life results exist to compare against. We therefore run simplified versions of each in one shared loop, with the same model, tasks and training budget, so that differences come from the judge, and from the retries only the non-\emph{oracle} judges get (3.1). This is a deliberate control rather than a substitute for the original systems.

\section{The creature}

\begin{figure}[htbp]
\centering
\includegraphics[width=0.9\linewidth]{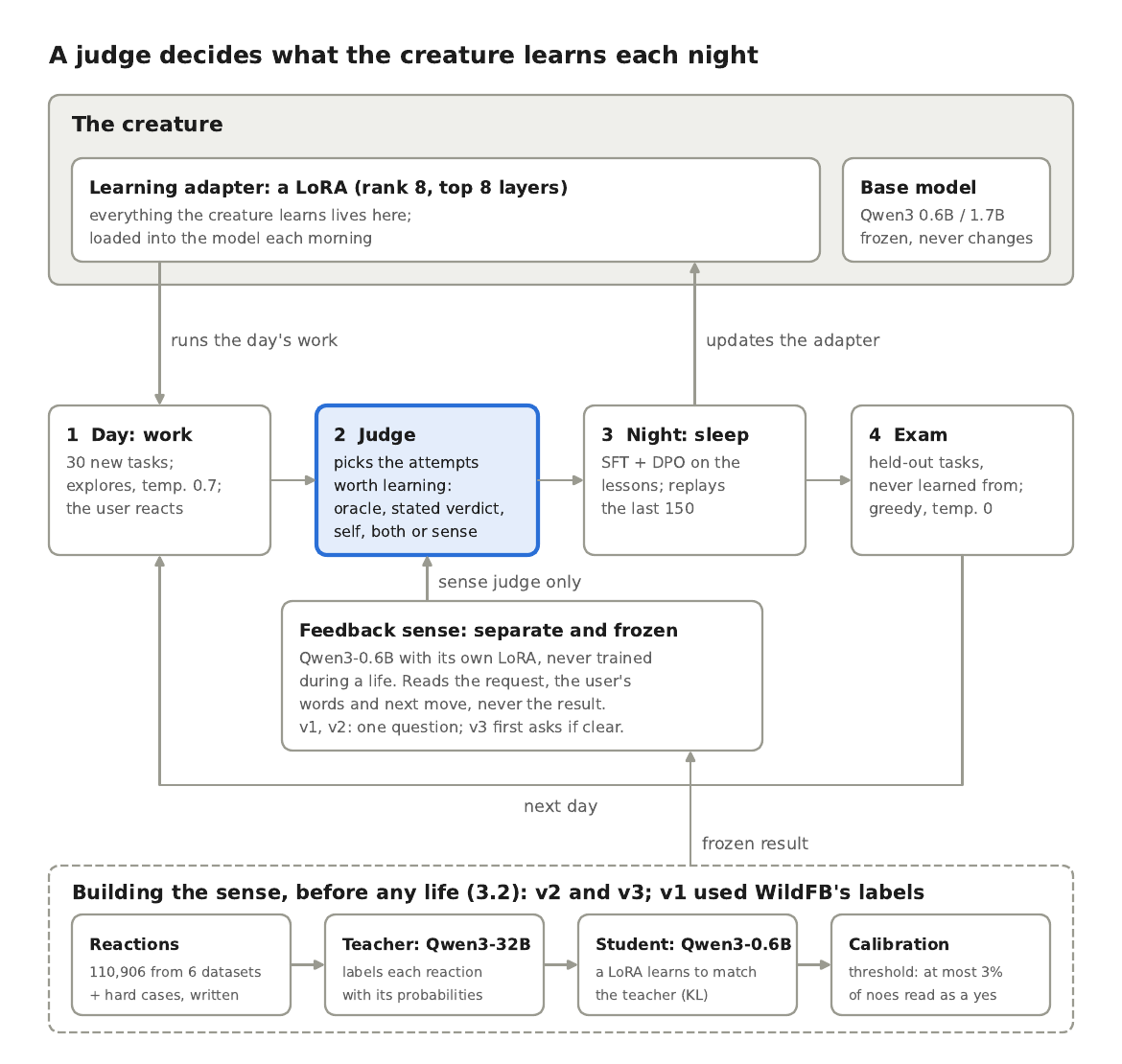}
\caption{The creature's parts (top), its daily cycle (middle), the feedback sense (bottom) and how the sense was built, once, before any life (dashed); the judge, highlighted, is what the experiments vary}
\label{fig:1}
\end{figure}

We call the system a creature: a small language-model agent that works by day and is fine-tuned at night (its ``sleep'') on lessons chosen by a judge. These names label an ordinary continual fine-tuning loop; they make no claim about biology. Its base model never changes; everything it learns lives in a small LoRA adapter, trained each night on the day's lessons (Fig.~\ref{fig:1}).

\subsection{The daily cycle}

\textbf{Body.} A frozen Qwen3 \cite{yang2025qwen3} base model (Fig.~\ref{fig:1}, top right), 0.6B or 1.7B parameters, served with vLLM \cite{kwon2023vllm} with thinking mode off. Each turn the agent replies in JSON with a short thought, an action and its input, as ``a personal assistant for one user''. The learning adapter (Fig.~\ref{fig:1}, top left) is a LoRA \cite{hu2022lora} of rank 8 on the attention and feed-forward projections of the top 8 layers, loaded into vLLM at run time so each morning's creature uses the previous night's adapter.

\textbf{Day.} In the first step of each day (Fig.~\ref{fig:1}, step 1), the creature works through a fixed number of new tasks (30 a day in every experiment), sampling at temperature 0.7 so it explores. After each attempt the simulated user reacts in words. The true outcome is logged for analysis, but the creature sees it only when the \emph{oracle} is its judge.

\textbf{Judge.} A judge then decides which attempts become lessons (Fig.~\ref{fig:1}, step 2, highlighted because it is the one part the experiments vary). We compare five (Table~\ref{tab:3}), each a simplified stand-in for a published family of methods.

\begin{longtable}[]{@{}
  >{\raggedright\arraybackslash}p{(\linewidth - 6\tabcolsep) * \real{0.2500}}
  >{\raggedright\arraybackslash}p{(\linewidth - 6\tabcolsep) * \real{0.2500}}
  >{\raggedright\arraybackslash}p{(\linewidth - 6\tabcolsep) * \real{0.2500}}
  >{\raggedright\arraybackslash}p{(\linewidth - 6\tabcolsep) * \real{0.2500}}@{}}
\caption{The judges. Only the sense condition uses the feedback sense.}\label{tab:3}\tabularnewline
\toprule\noalign{}
\begin{minipage}[b]{\linewidth}\raggedright
Judge
\end{minipage} & \begin{minipage}[b]{\linewidth}\raggedright
Who decides
\end{minipage} & \begin{minipage}[b]{\linewidth}\raggedright
What it sees
\end{minipage} & \begin{minipage}[b]{\linewidth}\raggedright
Stands in for
\end{minipage} \\
\midrule\noalign{}
\endfirsthead
\toprule\noalign{}
\begin{minipage}[b]{\linewidth}\raggedright
Judge
\end{minipage} & \begin{minipage}[b]{\linewidth}\raggedright
Who decides
\end{minipage} & \begin{minipage}[b]{\linewidth}\raggedright
What it sees
\end{minipage} & \begin{minipage}[b]{\linewidth}\raggedright
Stands in for
\end{minipage} \\
\midrule\noalign{}
\endhead
\bottomrule\noalign{}
\endlastfoot
\emph{oracle} & the benchmark's own score: reward 0.5 or more (WebShop), a full pass ($\tau^2$-bench) & the true outcome & training only on verified successes (STaR \cite{zelikman2022star}, rejection-sampling fine-tuning \cite{yuan2023scaling}); an upper bound no deployed system has \\
\emph{stated verdict} (``\emph{stated}'' in figures) & a fixed rule, no model & the kind of reaction the simulated user chose to give, not its words: praise or ``perfect'' is a yes; silence, settling or moving on is nothing; anything else is a no & learning from explicit user feedback, read without error; an ideal no deployed system has \\
\emph{self} & the creature itself, reflecting on its attempt & its own steps and what it bought; not the user's reaction & self-rewarding models \cite{yuan2024selfrewarding} \\
\emph{both} & the creature itself, reflecting on its attempt & its own steps, what it bought, and the user's words & Reflexion-style reflection with outside feedback \cite{shinn2023reflexion} \\
\emph{sense} & a separate, frozen model (3.2) & the request and the user's next move and words; never the result & our proposal \\
\end{longtable}

\emph{Self} and \emph{both} are the same model with the same prompt; only \emph{both} has the user's reaction in its prompt; \emph{self} is told the user said nothing. The \emph{stated verdict} is not a cruder sense but its ceiling: what a reader of reactions would achieve if it never misread one. With a user who says ``perfect'' only after a fully correct purchase (reward 1) and stays silent 20\% of the time, as our plain user does (Section 4), the \emph{stated verdict} accepts fewer attempts than the \emph{oracle} and never a failure. A last condition, \emph{none}, has no judge and no sleep: the model never learns.

Where the benchmark allows it, an attempt judged bad by any judge but the \emph{oracle} gets one retry, and a successful retry becomes a lesson and a preference pair (Section 4); the \emph{oracle} therefore yields no retries and no preference pairs. None of the judges is the original implementation of its method; each is the simplest version that fits the same nightly loop, so differences between conditions come from the judge, together with the retries and preference pairs that only the other judges produce.

\textbf{Night.} At night (Fig.~\ref{fig:1}, step 3) the adapter, and only the adapter, is trained on the lessons with supervised fine-tuning, plus DPO \cite{rafailov2023dpo} ($\beta$ = 0.1) on the preference pairs. Learning rate 1e-4, 2 epochs, at most 40 steps a night. Each night replays the 150 most recent training examples, one per step of a lesson, so old lessons fade out as new ones arrive; on WebShop a lesson takes about 3 or 4 steps, so where each lesson counts once the replay holds about 40--50 lessons, and on $\tau^2$-bench it holds the 150 most recent lessons. A lesson can count once for each of its steps, or once in all, with its loss spread across its steps so that a 15-step attempt counts as much as a 3-step one; Section 4 says which experiments used which.

\textbf{Exams.} The exam (Fig.~\ref{fig:1}, step 4) uses fixed held-out tasks, never learned from, attempted greedily (temperature 0): every morning on WebShop (less often late in the long lives; Section 4), every 3 days on $\tau^2$-bench, and on day 0 before any learning. The next day starts with the adapter that night produced (the loop along the bottom of Fig.~\ref{fig:1}), so each day's work uses everything learned so far.

\subsection{The feedback sense}

The sense (Fig.~\ref{fig:1}, bottom) is the creature's source of discernment, and is used only in the sense condition: a second, frozen Qwen3-0.6B with its own LoRA (rank 16), served beside the creature and never trained during a life. It reads three things: the user's request, the kind of move the user made next (replied, made a new request, asked to adjust, repeated the request, left, said nothing) and the user's words. It never sees what the creature did or how it turned out, so it reads people rather than judging the work. Because the sense is frozen and separate, the creature cannot change how it is judged.

We built three versions, one after another, each meant to fix a weakness of the one before. All three share the model, the inputs and the answers described next; they differ in what they learned from and in how they decide (Table~\ref{tab:4}). Section 4 says which experiment used which.

\emph{What every version shares.} The sense gives one of five answers: 100 (exactly right), 80 (acceptable, or good enough to build on), 20 (one part missed what was asked), 0 (wrong) or null (unclear); Table~\ref{tab:4} gives examples. They form two groups with a gap between them: 100 and 80 are a yes, 20 and 0 are a no, and there is no middle score. A reaction that does not clearly belong to one side is null, and nothing is learned from it. The gap is deliberate. The creature should learn only from a clear signal, so the sense must commit to a side or abstain, rather than hand the judge a lukewarm 50 to round one way or the other. Null comes about when the reaction says nothing about the result (silence, ``hmm'', settling, leaving without a word) or when the sense is not confident enough of its answer (under its confidence threshold); v3 adds a third way (below). Each answer begins with a different token (0, 2, 8, 1, n), so the sense decides in one forward pass, and its confidence is the probability of that first token.

A yes makes the attempt a lesson; an 80 counts half, since ``fine, next'' says less than praise. A no is not wasted on WebShop: it triggers the retry described in Section 4, with a hint matched to the answer, and if the sense reads the user's reaction to the retry as a yes, the retry becomes a lesson and a preference pair. On $\tau^2$-bench, which has no retry, a no teaches nothing.

\emph{Sense v1: one dataset's labels.} The first version learned without a teacher, from 45,514 reactions: 39,775 from WildFB \cite{peng2026wildreward}, whose own four-level satisfaction labels we mapped to 0, 20, 80 and 100 (rejection, correction, positive engagement, thanks or praise), and 5,739 silent or non-committal replies we added as null. It reads with a fixed confidence threshold of 0.70 instead of a calibrated one. On held-out WildFB reactions it gave a verdict on 66\%, 94\% of them on the right side, with 2.9\% false accepts. Its weakness shows on the reactions of Table~\ref{tab:4}: it took warm thanks for a failed attempt (``Thank you so much for trying! I'll look on my own'') and a lukewarm ``I guess that'll do'' for full praise, both with confidence above 0.9. One dataset's labels hold few such hard cases (politely unhappy, settling, sarcastic), and v2 set out to cover them.

\emph{Sense v2: a teacher.} We built v2 by distillation \cite{hinton2015distilling}, from conversations unrelated to our tests. Its raw material is 110,906 moments from six public datasets where an assistant delivered something and the user's next turn shows how they took it: WildFB \cite{peng2026wildreward} and WildFeedback \cite{shi2026wildfeedback} (30,000 each, both drawn from WildChat \cite{zhao2024wildchat}), Taskmaster 1-3 \cite{byrne2019taskmaster} (20,000), MultiWOZ 2.2 \cite{zang2020multiwoz} and ReDial \cite{li2018redial} (15,000 each) and CCPE-M \cite{radlinski2019ccpe} (906; Table~\ref{tab:A3}). Turns where the assistant only asked a question were skipped, since the answer is not a reaction, and shopping was left out, since it is the test domain. Real data rarely holds the hard cases, so the teacher also wrote 4,728 reactions of eight kinds (politely unhappy, settling, vague, sarcastic, ``not my style'', firm rejection, building on the result, plain acceptance) across ten task types and six personas, chosen to differ from our WebShop user's.

A teacher, Qwen3-32B, labelled every reaction under a written guide. It saw the assistant's reply as context; the student never does. Instead of one label we kept the teacher's probabilities over the five answers, read from the log-probabilities of its first answer token, so the student learns the teacher's doubt as well as its choice. On the two datasets that carry their own signal of success, the teacher agreed with it on yes or no 81\% (WildFB) and 85\% (WildFeedback) of the time, and it called about half of the WildFB reactions unclear. The student inherits the teacher's mistakes as well as its knowledge; the confidence threshold below is meant to keep them out of the creature's lessons.

\begin{figure}[htbp]
\centering
\includegraphics[width=0.75\linewidth]{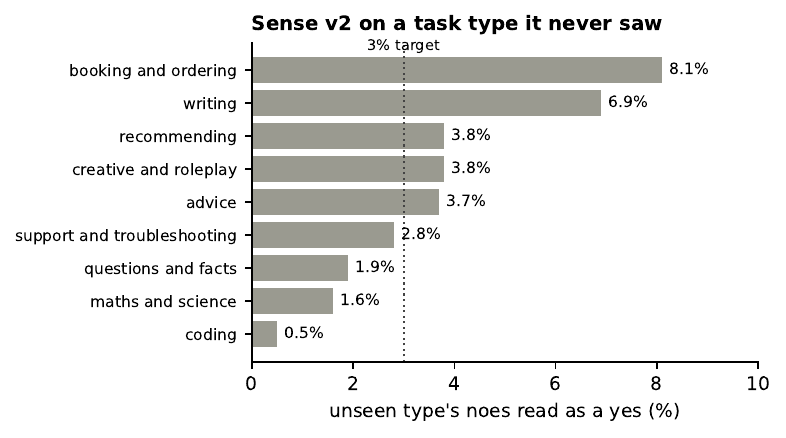}
\caption{Sense v2 trained nine more times, each time without one task type and tested on it: the share of that type's noes it read as a yes (leave-one-task-out builds)}
\label{fig:8b}
\end{figure}

The student learns to match the teacher's probabilities (KL divergence) on 47,800 reactions, balanced so that no task type or answer dominates, for one epoch (learning rate 1e-4, batch 16, under an hour on one GPU). In half the training rows the request is hidden, so the sense must rely on the reaction itself. After training, a temperature is fitted on 2,000 held-out reactions, and a confidence threshold is chosen as the lowest that keeps false accepts (a no read as a yes) at or under 3\%; below it, the sense answers null. This is sense v2. Its threshold is 0.65. On the held-out reactions it gave a verdict on 30\% of them, agreed with the teacher's side on 92\% of those, and read 2.9\% of the teacher's noes as a yes. Trained nine more times, each time without one task type and tested on it, it read 0.5-8.1\% of the unseen type's noes as a yes (Fig.~\ref{fig:8b}). Four types stayed under the 3\% target and three went just over it (3.7-3.8\%); writing (6.9\%) and booking and ordering (8.1\%), the type closest to shopping, went well over. A new kind of task can loosen the sense. And v2 still asks a single question, which score, so a reaction must look like a no to be refused: warm thanks for the effort after a failed attempt carries no complaint and can score as approval.

\begin{figure}[htbp]
\centering
\includegraphics[width=1.0\linewidth]{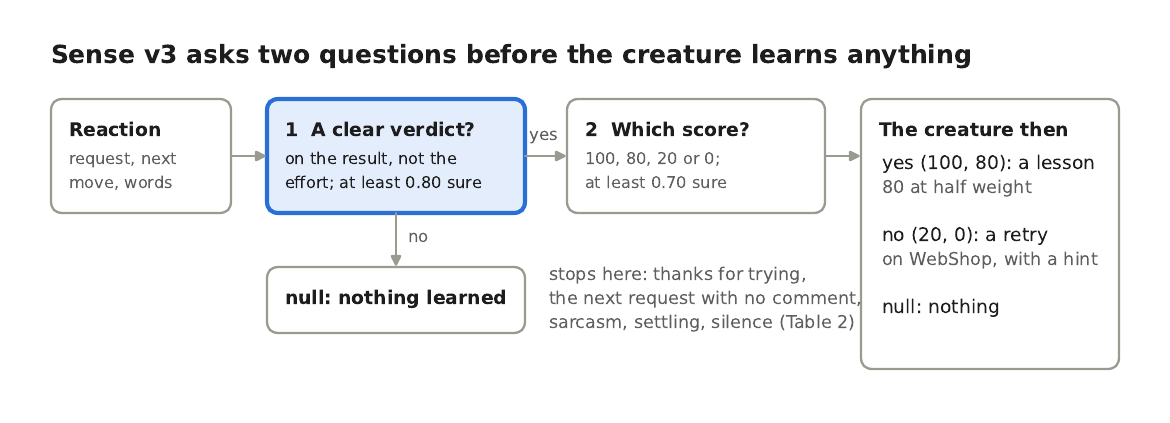}
\caption{How sense v3 decides: a reaction must pass both gates before the creature learns from it. v1 and v2 ask only the second question, with one threshold (0.70 and 0.65)}
\label{fig:9}
\end{figure}

\begin{figure}[htbp]
\centering
\includegraphics[width=0.75\linewidth]{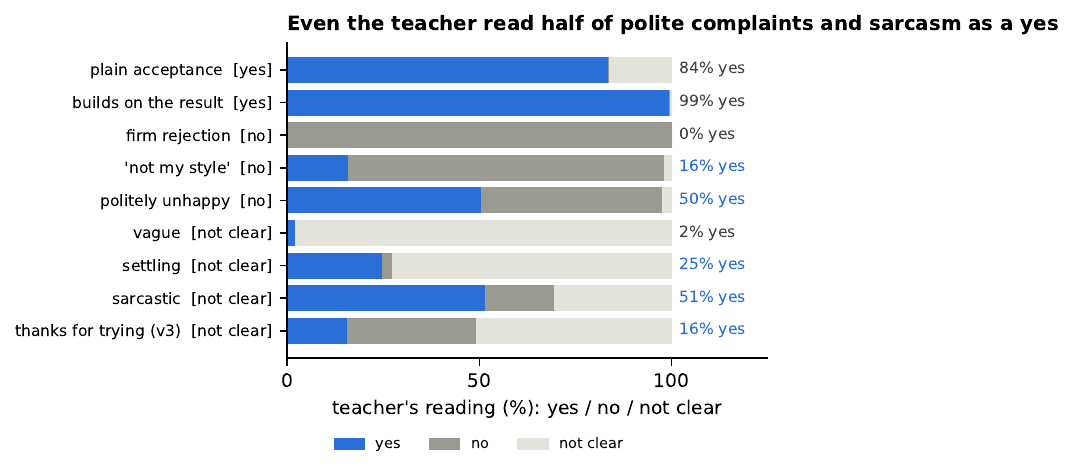}
\caption{How the teacher, under v3's guide, read the 6,506 reactions it wrote; in brackets, the reading each kind was written to have}
\label{fig:8a}
\end{figure}

\emph{Sense v3: two questions.} It replaces v2's single question with two, so that thanks for the effort can be told from approval of the result. First, is the reaction a clear verdict on the result at all? Thanks for trying, sarcasm, hedging, settling, silence and moving on without a word are not. Only if it is: which score? The teacher answered both, under a revised guide, for 63,248 reactions: up to 2,000 for each pairing of task type and v2 answer, the eight written kinds again, and 1,778 new reactions of a ninth kind, warm thanks for the effort after a failure. It judged 41\% of them not clear. The teacher is not a perfect reader either (Fig.~\ref{fig:8a}): of the reactions it wrote itself, it read half of the politely unhappy and half of the sarcastic ones as a yes, a quarter of the settling ones, and 16\% of the warm thanks for trying. The student inherits these mistakes; the first question and the thresholds below are what is meant to keep them out of the creature's lessons. One LoRA learns both answers from 44,746 of them, the score weighted by how clear the teacher found the reaction. Its two thresholds, 0.80 for a clear verdict and 0.70 for the score, were chosen on 2,000 held-out reactions as the one that keeps false accepts, now also counting an unclear reaction read as a yes, at 3\% while keeping the most real yeses. On those reactions it kept 44\% of the teacher's yeses and agreed with the teacher's side on 95\% of the verdicts it gave. In use, a reaction must pass both gates before the creature learns from it (Fig.~\ref{fig:9}). The first gate changes how some common reactions are read (Table~\ref{tab:4}): v2's guide scored moving on to the next request without a word as an 80, a half-weight yes (in use, v2 mostly read it as unclear), and thanks for trying could pass as approval; for v3, neither is a clear verdict.

\emph{Side by side.} Table~\ref{tab:4} shows how the three versions read the same reactions.

\begin{longtable}[]{@{}
  >{\raggedright\arraybackslash}p{(\linewidth - 10\tabcolsep) * \real{0.2600}}
  >{\raggedright\arraybackslash}p{(\linewidth - 10\tabcolsep) * \real{0.1000}}
  >{\raggedright\arraybackslash}p{(\linewidth - 10\tabcolsep) * \real{0.1300}}
  >{\raggedright\arraybackslash}p{(\linewidth - 10\tabcolsep) * \real{0.1100}}
  >{\raggedright\arraybackslash}p{(\linewidth - 10\tabcolsep) * \real{0.0600}}
  >{\raggedright\arraybackslash}p{(\linewidth - 10\tabcolsep) * \real{0.3400}}@{}}
\caption{How the three versions read the same reactions, and what the creature does. The v2 and v3 columns are examples from their labelling guides; v3 first asks whether the reaction is a clear verdict, and scores it only if it is. v1 had no guide, so its column is v1's own reading of each reaction, tried under three different requests; ``80 or null'' means it varied with the request.}\label{tab:4}\tabularnewline
\toprule\noalign{}
\begin{minipage}[b]{\linewidth}\raggedright
Reaction
\end{minipage} & \begin{minipage}[b]{\linewidth}\raggedright
v1
\end{minipage} & \begin{minipage}[b]{\linewidth}\raggedright
v2
\end{minipage} & \begin{minipage}[b]{\linewidth}\raggedright
v3: clear verdict?
\end{minipage} & \begin{minipage}[b]{\linewidth}\raggedright
v3
\end{minipage} & \begin{minipage}[b]{\linewidth}\raggedright
What the creature does
\end{minipage} \\
\midrule\noalign{}
\endfirsthead
\toprule\noalign{}
\begin{minipage}[b]{\linewidth}\raggedright
Reaction
\end{minipage} & \begin{minipage}[b]{\linewidth}\raggedright
v1
\end{minipage} & \begin{minipage}[b]{\linewidth}\raggedright
v2
\end{minipage} & \begin{minipage}[b]{\linewidth}\raggedright
v3: clear verdict?
\end{minipage} & \begin{minipage}[b]{\linewidth}\raggedright
v3
\end{minipage} & \begin{minipage}[b]{\linewidth}\raggedright
What the creature does
\end{minipage} \\
\midrule\noalign{}
\endhead
\bottomrule\noalign{}
\endlastfoot
``Perfect, exactly what I needed'' & 100 & 100 & yes & 100 & learns it as a lesson, full weight \\
``Ok, that works'' & 100 & 80 & yes & 80 & learns it as a lesson: full weight for v1, half for v2 and v3 \\
``Great, now can you also\ldots{}'' (building on it) & 80 or null & 80 & yes & 80 & learns it as a lesson, half weight (v1: only sometimes) \\
the next request, with no comment & 80 or null & 80 (in use, mostly null) & no & null & v1: sometimes learns at half weight; v2: by its guide half weight, in use mostly nothing; v3: nothing \\
``Thank you so much for trying! I'll look on my own'' & 100 & often a yes & no & null & v1: learns a failure at full weight; v2: often does; v3: nothing \\
``I guess that'll do''; ``hmm''; silence & 100; null; null & null & no & null & v1 learns ``I guess that'll do'' at full weight; otherwise nothing \\
``I said size 9'' & 20 & 20 & yes & 20 & retries, told to change only what the user asked for \\
``No, that's not what I asked''; repeats the request & 0; null & 0 & yes & 0 & retries, told to start again and read the request carefully (v1: only after the explicit no) \\
\end{longtable}

\section{Experimental setup}

We ran five experiments on two benchmarks, each with an \emph{oracle} condition as the upper bound and several seeds per condition (Table~\ref{tab:1}). Experiment 1 has three parts: 1-A compares five judges over 14 days, 1-B follows two of them for a year, and 1-C tries four adapter designs. The creature, its nightly training and the judges are described in Section 3, where each judge stands in for a published family of methods (Table~\ref{tab:3}); Table~\ref{tab:A1} lists every setting.

\textbf{Why small models.} Learning on the job matters most where a model lives with its user: on a phone, a laptop or a local server, where the user's reactions need never leave the device. Models there are small, so we study Qwen3 at 0.6B (all experiments) and 1.7B (Experiment 5). The whole loop fits on one 24 GB GPU: the sense is itself a 0.6B model served beside the creature, and a night's training on WebShop takes minutes (see Compute below). Small models also leave room to improve (the 0.6B starts at 13 of 100 on WebShop), and their low cost let us run 127 lives, with three to five seeds for most conditions. Whether larger models, which may judge themselves better, decay less when they are the judge is left open (Section 7).

\textbf{Benchmarks.} WebShop \cite{yao2022webshop} (Experiments 1-4) is a simulated online shop built from 1.18 million real products, with 12,087 shopping requests written by crowd workers. We used its small setting: 1,000 products, 6,910 shopping requests generated from their attributes, and the 13 requests written by people that fall within those products. The creature reads a request, searches, browses and buys; an attempt ends with a purchase or after 15 steps. The reward, 0 to 1, scores how well the bought product matches the request on its type, attributes, options and price \cite{yao2022webshop}. $\tau^2$-bench \cite{barres2026tau2} (Experiment 5), which extends $\tau$-bench \cite{yao2025taubench}, is a customer-service benchmark: the agent talks with a simulated customer and calls the domain's tools to change bookings, orders or phone plans under a written policy. Success is judged from the final state of the database. In telecom both sides act: the customer also uses tools on their own phone, as the agent directs \cite{barres2026tau2}. We used the train and test splits published in $\tau^2$-bench's repository and its airline, retail and telecom domains, with Gemini 3 Flash as the customer and as $\tau^2$-bench's graders in place of the published GPT models. Scores are therefore comparable to published ones, not identical.

\textbf{Simulated users.} We wrote WebShop's simulated user. In Experiments 1 and 2 it reacts plainly to the outcome: it stays silent 20\% of the time and otherwise says one fixed kind of sentence per outcome (no purchase, wrong kind of product, close but not quite right, perfect). From Experiment 3 on it reacts as people do: its behaviour is drawn from a mix tied to the outcome and to a persona (clear thanks, complaints, polite give-ups, vague replies, moving on without a word), and Gemini words each reaction (Table~\ref{tab:A2}; prompt in Appendix A). Because we designed this user, results that depend on it carry that design, and the sense's live test (5.4) uses its reactions. Two things limit the risk: the sense was never trained on shopping reactions, and 5.5 tests it against a simulated customer we did not design ($\tau^2$-bench's).

\textbf{Which sense.} The \emph{sense} judge used a different version of the feedback sense (3.2) in each step of the work, frozen in every life (Table~\ref{tab:1}, column \emph{Sense}): Experiment 1 has no sense judge; Experiment 2 uses sense v1, which learned one dataset's own labels; Experiment 3 uses sense v2, which learned a teacher's labels; Experiments 4 and 5 use sense v3, which first asks whether a reaction is a clear verdict. None of the versions trained on shopping or on our simulated users' words. But v3 is not blind to the WebShop user's failure: it was built after Experiment 3, its ninth kind of reaction was added because Experiment 3 exposed that failure, its labelling guide (src/lab/sense/guide3.md) includes, as a worked example, a reaction almost word for word like the give-up of Example 1 (``That's perfectly fine, thank you so much for trying! I'll look on my own''), and its test on recorded reactions (5.4, Fig.~\ref{fig:3}) uses that user's words.

\textbf{Protocol.} The first 100 of WebShop's synthetic goals are held out from practice; the exam uses 50 of them, plus 13 written by people, every morning, starting on day 0 before any learning. The longest lives were examined less often later on: the year-long lives daily to day 30, every second day to day 90 and weekly after, and Experiment 4 every second day after day 30. With the \emph{oracle} as judge, an attempt with reward 0.5 or more becomes a lesson; we call an attempt below 0.5 a failure throughout. An attempt judged bad by any judge other than the \emph{oracle} gets one retry, with the user's comment and the creature's own lesson as a hint (Appendix A); a successful retry becomes a lesson (without the hint) and a preference pair (the retry over the failed attempt). On $\tau^2$-bench the creature practises 30 conversations a day on the airline and retail training tasks; the exam, on day 0 and every third day after, uses the airline (20) and retail (40) test tasks and telecom (40), a domain it never practises. $\tau^2$-bench has no retry: the customer cannot be asked to react twice. In sleep, each step of a lesson counts as a lesson of its own in Experiments 1-3, and each lesson counts once in all in the sleep variant of Experiment 3 and in Experiments 4 and 5 (3.1, Table~\ref{tab:1}). A seed fixes which training tasks a life works on and in what order, and the simulated user's random draws (silence, wording, persona and behaviour), so lives with the same seed under different judges start from the same tasks and the same user; the exam tasks are the same for every life. Practice sampling is not seeded, so even such lives differ from their first attempt.

\begin{longtable}[]{@{}
  >{\raggedright\arraybackslash}p{(\linewidth - 10\tabcolsep) * \real{0.0700}}
  >{\raggedright\arraybackslash}p{(\linewidth - 10\tabcolsep) * \real{0.2300}}
  >{\raggedright\arraybackslash}p{(\linewidth - 10\tabcolsep) * \real{0.1500}}
  >{\raggedright\arraybackslash}p{(\linewidth - 10\tabcolsep) * \real{0.2600}}
  >{\raggedright\arraybackslash}p{(\linewidth - 10\tabcolsep) * \real{0.0800}}
  >{\raggedright\arraybackslash}p{(\linewidth - 10\tabcolsep) * \real{0.2100}}@{}}
\caption{The experiments. \emph{Sense}: the version of the feedback sense (3.2) that the \emph{sense} judge used; -- : no sense judge.}\label{tab:1}\tabularnewline
\toprule\noalign{}
\begin{minipage}[b]{\linewidth}\raggedright
Exp.
\end{minipage} & \begin{minipage}[b]{\linewidth}\raggedright
Question
\end{minipage} & \begin{minipage}[b]{\linewidth}\raggedright
Benchmark, model
\end{minipage} & \begin{minipage}[b]{\linewidth}\raggedright
Judges
\end{minipage} & \begin{minipage}[b]{\linewidth}\raggedright
Sense
\end{minipage} & \begin{minipage}[b]{\linewidth}\raggedright
Seeds $\times$ days
\end{minipage} \\
\midrule\noalign{}
\endfirsthead
\toprule\noalign{}
\begin{minipage}[b]{\linewidth}\raggedright
Exp.
\end{minipage} & \begin{minipage}[b]{\linewidth}\raggedright
Question
\end{minipage} & \begin{minipage}[b]{\linewidth}\raggedright
Benchmark, model
\end{minipage} & \begin{minipage}[b]{\linewidth}\raggedright
Judges
\end{minipage} & \begin{minipage}[b]{\linewidth}\raggedright
Sense
\end{minipage} & \begin{minipage}[b]{\linewidth}\raggedright
Seeds $\times$ days
\end{minipage} \\
\midrule\noalign{}
\endhead
\bottomrule\noalign{}
\endlastfoot
1-A & Who should decide what is learned? & WebShop, 0.6B & \emph{oracle}, \emph{self}, \emph{both}, \emph{stated verdict}, \emph{none} & -- & $5 \times 14$ (pilot $1 \times 6$) \\ \addlinespace
1-B & Does the decline pass with time? & WebShop, 0.6B & \emph{both}, \emph{oracle} & -- & $5 \times 365$ (\emph{oracle} $1 \times 365$) \\ \addlinespace
1-C & Does it depend on the adapter? & WebShop, 0.6B & \emph{both} with four adapter designs & -- & $5 \times 14$ each \\ \addlinespace
2 & Does a separate, frozen judge help? & WebShop, 0.6B & \emph{oracle}, \emph{both}, \emph{sense} & v1 & $5 \times 14$ \\ \addlinespace
3 & Does it survive realistic users? & WebShop, 0.6B, realistic user & \emph{oracle}, \emph{both}, \emph{sense}; then the same with each lesson counted once in sleep and sleep only after 5 new lessons & v2 & $5 \times 14$, twice \\ \addlinespace
4 & Does waiting for a clear verdict help? & WebShop, 0.6B, realistic user & \emph{oracle}, \emph{both}, \emph{sense}; each lesson counted once in sleep & v3 & $5 \times 60$ (planned as 14, then extended) \\ \addlinespace
5 & What if the user cannot see the outcome? & $\tau^2$-bench, 0.6B and 1.7B & \emph{oracle}, \emph{sense}; each lesson counted once in sleep & v3 & $3 \times 30$ per size \\
\end{longtable}

\textbf{Measures.} The exam score over days, as the mean of seeds with the lowest and highest life. On WebShop it is the mean reward, 0 to 100, on the 50 synthetic tasks (the 13 written by people were also run each morning but are not used in the results); on $\tau^2$-bench it is the share of tasks fully done, per domain. We also report the number of lessons a day; the share of lessons whose attempt had truly succeeded; and, for the sense, its accuracy on 302 fixed recorded reactions. Statistics treat each life as one sample; two groups are compared with a Welch t-test and a Mann-Whitney U test, and the primary claims are Holm-corrected \cite{holm1979}. Appendix B lists every test. The data, the code of the creature and the scripts that recompute every statistic and figure (src/stats.py, src/figures.py) are public at \url{https://github.com/kratai-desci/research/tree/discernment/discernment}.

\textbf{Compute.} Lives ran on single NVIDIA L4 GPUs on Google Cloud, eight lives at a time on one GPU in Experiment 1 and two at a time in the others. In all, 127 lives ran for about 1,300 hours and made about 351,000 attempts: practice, retries and exam tasks together. On WebShop a night's training took 1-4 minutes (median 2, the 0.6B \emph{oracle} life of Experiment 4). Building the three senses took about 8 GPU-hours (v1 1.5; v2 3.6 and v3 3.1, teacher labels and student training), and the nine leave-one-task-out builds of v2 another 5.6.

\section{Results}

On WebShop the judge decided each life. With the true outcome as judge, the creature rose from 13 to 62 within a week and held there; self-judging rose and then fell, \emph{sense v2} collapsed when it let polite give-ups in, and \emph{sense v3}, which waits for a clear verdict, came close to the \emph{oracle} for two weeks before fading. On $\tau^2$-bench, where the customer cannot see whether a task was done, learning from reactions collapsed and even the \emph{oracle} lowered the overall score (5.5). Across the WebShop lives, the share of failed attempts among the lessons tracked the outcome closely (5.6).

Each subsection is one experiment, in the order they were run: Experiment 1 asks who should judge before there is any sense (5.1); Experiments 2-4 try the three versions of the sense in turn (5.2-5.4); Experiment 5 takes sense v3 to a benchmark whose customer cannot see the outcome (5.5); and 5.6 looks across all WebShop lives. WebShop scores are the exam: the mean reward, 0 to 100, on 50 held-out tasks never learned from. Each number is the mean of 5 seeds with the lowest and highest life in brackets.

\textbf{What a WebShop life learns.} Nearly all of what is learned, and lost, on WebShop is to finish the purchase. Before any learning the creature bought something on only 19\% of the exam tasks and ran into the 15-step limit on about 80\% (day 0, all 115 WebShop lives). With the \emph{oracle} as judge it bought on 99\% by day 14, in 3 steps on average (25 lives). The reward of a purchase hardly changed: 67 out of 100 before, 63 after. The 12 lives that collapsed (exam of 3 or below on day 14) bought on 4\%, below where they began. So ``collapse'' on WebShop means the creature stopped buying, and what the lessons teach is mostly to press \emph{Buy Now}, not to match products better. These counts come from the step logs in the level-3 archive, not from \texttt{data/}.

\subsection{Experiment 1: Before discernment}

\textbf{1-A: five judges.} The first part ran five lives that differed only in the judge, 5 seeds each for 14 days: the \emph{oracle} (the true reward), the user's \emph{stated verdict}, the creature's own reflection (\emph{self}), its reflection while hearing the user (\emph{both}), and no sleep at all (\emph{none}).

\begin{figure}[htbp]
\centering
\includegraphics[width=1.0\linewidth]{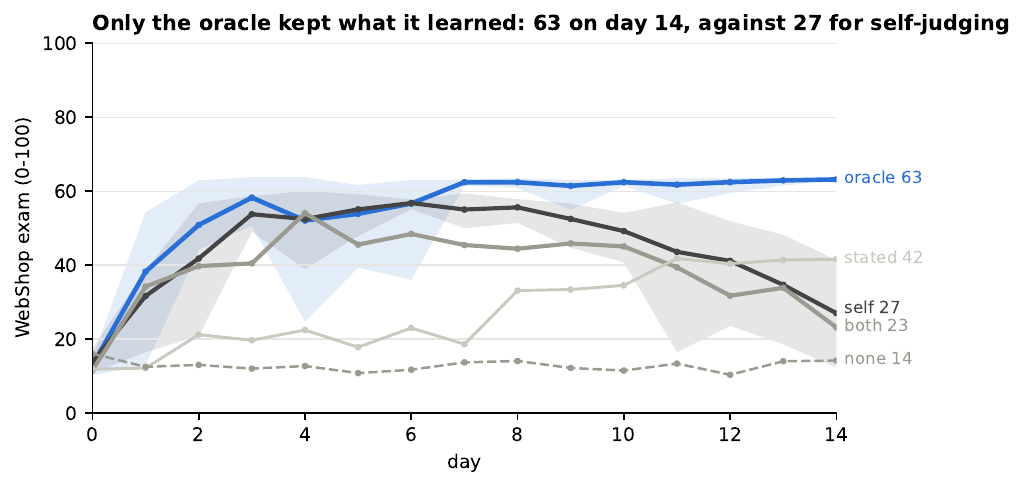}
\caption{Experiment 1-A, 5 seeds per judge, daily exams; one self life crashed early and is left out}
\label{fig:2}
\end{figure}

On day 14 the \emph{oracle} averaged 63.2 $\pm$ 0.2 (mean $\pm$ standard deviation over seeds), \emph{self} 27.0 $\pm$ 11.9, \emph{both} 23.1 $\pm$ 17.0, \emph{stated verdict} 41.6 $\pm$ 23.1 and \emph{none} 14.2 $\pm$ 2.2. The \emph{stated verdict} learned from only 0-7 lessons per life, none of them a failure; three of its lives reached 54-61 and two stayed at 14-19. The \emph{oracle} beat \emph{self} and \emph{both} clearly (Welch t-test, p = 0.009 and 0.006); with 4-5 seeds, self-judging did not end reliably above no learning at all (p = 0.12).

Self-judging learned as fast as the \emph{oracle} at first and then lost most of it. By day 7 the \emph{self} lives averaged 55 and the \emph{both} lives 45 (lives 32-59); by day 14 every one of those nine lives had fallen below its own day-7 score, one to 0. The \emph{oracle} reached 62 by day 7 and stayed there. The \emph{stated verdict} rose slowly and unevenly (14-61 on day 14). Only four \emph{self} lives completed the 14 days; the fifth crashed early and is left out. A 6-day pilot showed the same shape: the \emph{self} life rose from 13 to 34 by day 1 and fell to 9 by day 6.

\textbf{1-B: a year.} Does the decline pass with time? We let the \emph{both} creature, the self-judge that also hears the user, live for a year: 365 days, 5 seeds, beside one \emph{oracle} life. WebShop's practice tasks run out after about 227 days, so later days revisit old tasks.

\begin{figure}[htbp]
\centering
\includegraphics[width=1.0\linewidth]{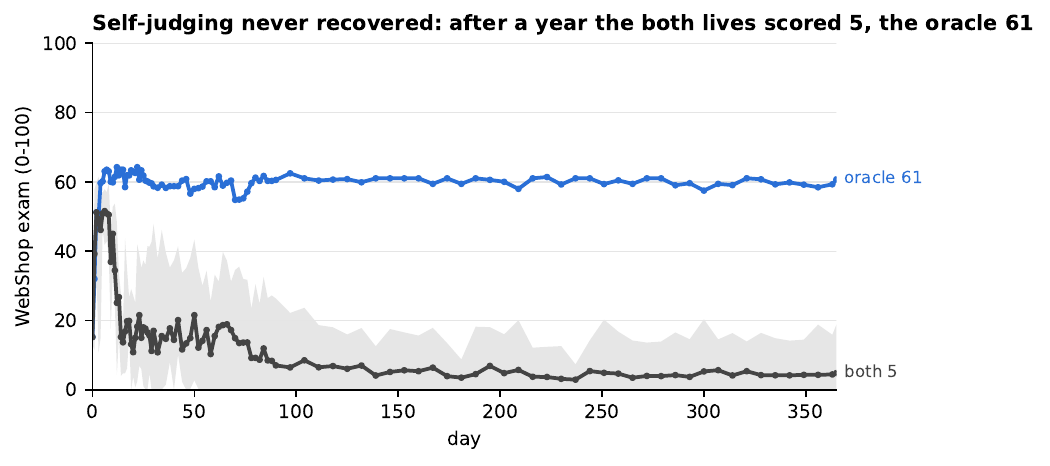}
\caption{Experiment 1-B, a year: 5 both lives and 1 oracle life; exams daily to day 30, every 2 days to day 90, then weekly}
\label{fig:7}
\end{figure}

It never recovered. The \emph{both} lives peaked around day 6 (52, lives 42-58), were back at their starting level of 15 by day 14, and ended the year at 5 (0-19), three of the five at 0. The \emph{oracle} life rose from 15 to 64 by day 14 and stayed between 55 and 64 for the rest of the year (61 on day 365). These are separate lives from the 14-day \emph{both} condition above (23 on day 14, lives 0-46); the two ranges overlap. A first attempt at this run crashed when two nights ran out of GPU memory at once and was restarted from scratch.

\textbf{1-C: four adapters.} Is it the adapter? If the adapter simply overwrote what it had learned, an adapter built to protect old learning might stop the decline. We gave the \emph{both} creature four designs, 5 seeds each for 14 days: one LoRA that keeps growing (as everywhere else), a LoRA whose learning rate halves every 4 days, a new rank-8 LoRA every 2 days with the older ones frozen, and the same with each new one smaller (rank 8 down to 1). All four stood at 46-53 on day 7, having peaked a few days earlier, and fell to 26-38 by day 14; the best of them still lost more than a quarter of its peak.

Neither time nor the adapter explains the decline. What the judge lets in does, and the rest of this section follows that thread.

\subsection{Experiment 2: Sense v1, a separate, frozen judge}

Experiment 2 gave the creature a judge it cannot change: the feedback sense in its first build (v1; 3.2), a frozen Qwen3-0.6B with its own LoRA that reads the user's reaction to each attempt. It ran against the \emph{oracle} and \emph{both}, 5 seeds, 14 days, with the same user as Experiment 1.

\begin{figure}[htbp]
\centering
\includegraphics[width=1.0\linewidth]{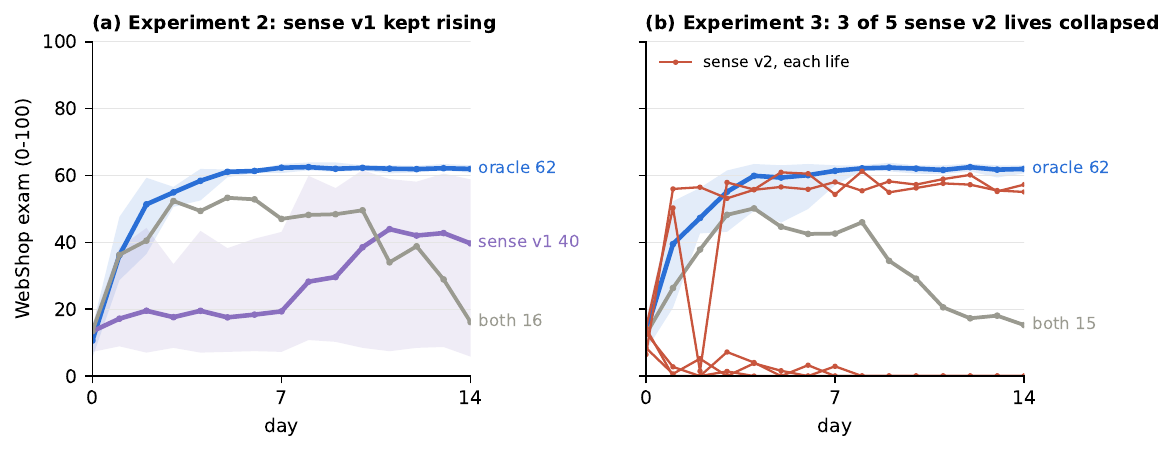}
\caption{Experiments 2 and 3, 5 seeds per judge, daily exams: mean of the lives, with the lowest-to-highest range shaded; in (b) each \emph{sense v2} life is drawn on its own}
\label{fig:10}
\end{figure}

\emph{Sense v1} learned more slowly but did not fall back (Fig.~\ref{fig:10}a). It also learned from very little: 0-7 lessons per life in 14 days, against 247-274 for the \emph{oracle}. By day 14, four of its five lives stood at 34-59, while every \emph{both} life had fallen below its day-7 score. On day 14 \emph{sense v1} (39.7 $\pm$ 21.3) was ahead of \emph{both} (16.1 $\pm$ 11.1) and behind the \emph{oracle} (62.0 $\pm$ 1.0). With 5 seeds and one \emph{sense v1} life at 6, the t-test finds neither gap significant (p = 0.07 and 0.08); a rank test, less swayed by that one life, puts the gap to the \emph{oracle} at p = 0.01 (Appendix B). Sense v1's own accuracy stayed fixed, as intended: on 302 recorded reactions it read 82.6\% correctly on day 0 and 82.8\% on day 14. A variant that read the sense through the creature's learning brain was planned, but its lives scored 0 from day 0 because of a setup fault and were stopped; it is not reported.

\subsection{Experiment 3: Sense v2 and realistic users}

Experiment 3 replaced the user with one that reacts as people do. Its behaviour is drawn from a mix tied to the outcome and a persona, and worded by Gemini: clear thanks, complaints, but also polite give-ups, vague replies and moving on without a word. \emph{Oracle}, \emph{both} and \emph{sense v2}, 5 seeds, 14 days.

The \emph{sense v2} lives split in two (Fig.~\ref{fig:10}b). Two reached 55 and 57 by day 14, close to the \emph{oracle}'s 62. The other three fell to 0-3 on day 1 and never recovered. Each had learned on night 1 from one to three polite give-ups that sense v2 had read as success (Example 1 below); 96-100\% of everything those three lives learned were failed attempts. The \emph{both} lives rose to 48 by day 3 and fell to 15 (2-36). On day 14 the \emph{sense v2} lives averaged 22.5 $\pm$ 30.8 against the \emph{oracle}'s 62.0 $\pm$ 1.1 (Welch t-test p = 0.045, Mann-Whitney p = 0.01); the spread is the split itself.

\textbf{Example 1: a give-up sense v2 took for success (Experiment 3, night 1).} The shopper asked for men's loafers in size 6.5 under \$60. The creature bought nothing (reward 0). The user ended:

\begin{quote}
``That's perfectly fine, thank you so much for trying to help me! I think I'll just leave it for now and head out.''
\end{quote}

Sense v2 scored this reaction 100, the top score, with confidence 0.92, and the attempt became a lesson. All five practice attempts that became bad lessons on night 1 in the three collapsed lives look like this: warm thanks for the effort, then the user leaves.

\textbf{The sleep variant.} Changing sleep did not remove the cause. When each lesson counted once, however many steps it took, and the creature slept only after 5 lessons, fewer lives collapsed, but 2 of 5 still ended at 0-3. The fault was in the judge, not the sleep.

\subsection{Experiment 4: Sense v3, waiting for a clear verdict}

Experiment 4 used sense v3, which waits for a clear verdict before it scores a reaction (3.2). Same user as Experiment 3, with each lesson counted once in sleep (Table~\ref{tab:1}).

\begin{figure}[htbp]
\centering
\includegraphics[width=1.0\linewidth]{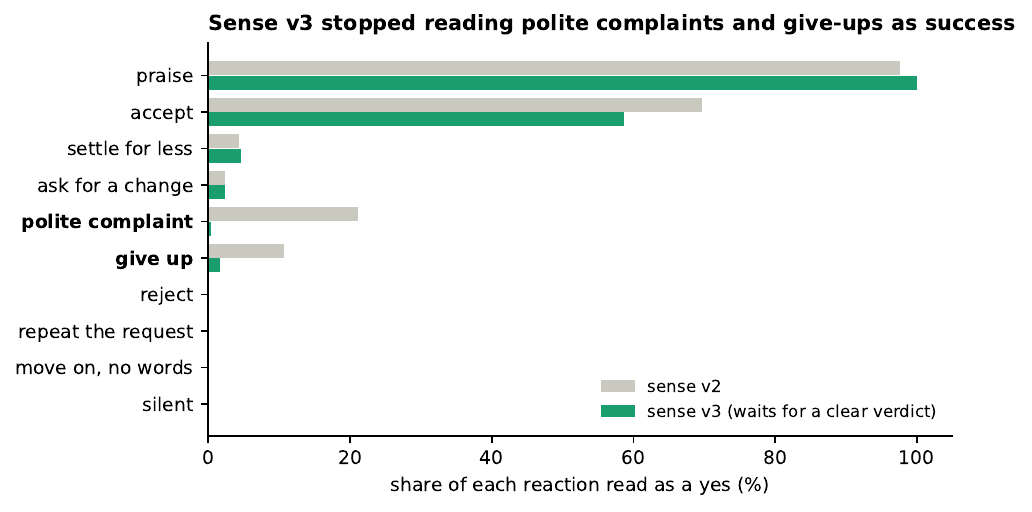}
\caption{The live test: sense v2 and v3 reading the same 2,622 reactions recorded in the Experiment 3 lives, up to 300 per behaviour}
\label{fig:3}
\end{figure}

\textbf{The live test.} Before the lives began, we had both versions read the same 2,622 reactions recorded in the Experiment 3 lives, up to 300 of each behaviour (Fig.~\ref{fig:3}). Sense v3 read 1.2\% of reactions that were neither praise nor acceptance as a yes, against 5.0\% for v2, and read 1.0\% of give-ups and polite complaints as success, against 15.8\% (Fisher exact test, p \textless{} 0.001). The price was a stricter reading of real approval: it took 59\% of plain acceptances as a yes, against 70\% for v2.

\textbf{The first 14 days.} Four of the five \emph{sense v3} lives stood at 53-60 on day 14, against the \emph{oracle}'s 63 (62-63). The fifth fell to 0 after learning from a single give-up on night 1. Averaged over all five lives \emph{sense v3} (45.1 $\pm$ 25.3) was not reliably below the \emph{oracle} by a t-test (62.8 $\pm$ 0.6; p = 0.19), because the one collapsed life dominates the spread; by a rank test it was (p = 0.01), since every \emph{sense v3} life scored below every \emph{oracle} life. The \emph{both} lives peaked at 57 on day 4 and fell to 19 by day 14, as before.

\textbf{Days 15-60: clean but scarce lessons fade.} Every life of Experiment 4 was extended to 60 days.

\begin{figure}[htbp]
\centering
\includegraphics[width=1.0\linewidth]{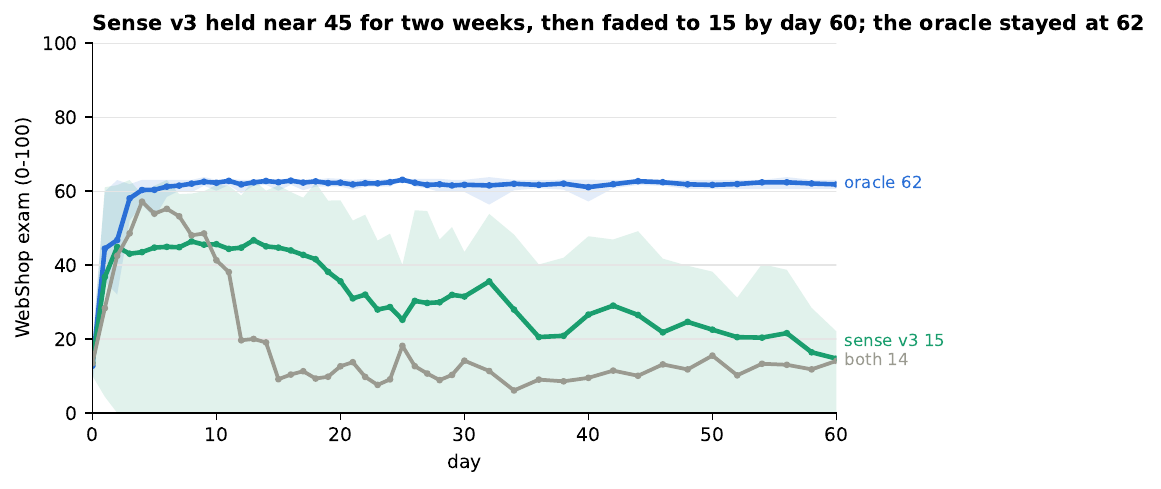}
\caption{Experiment 4, 5 seeds per judge, 60 days; exams daily to day 30, then every 2 days}
\label{fig:4}
\end{figure}

Life by life, the four \emph{sense v3} lives that did not collapse stood at 53-60 on day 14, 36-44 on day 30 and 9-22 on day 60; the fifth stayed at 0 throughout. On day 60 the \emph{oracle} (61.8 $\pm$ 0.9) was far above \emph{sense v3} (14.7 $\pm$ 9.7; Welch t-test p \textless{} 0.001).

The four \emph{sense v3} lives declined steadily after day 14, although what they learned stayed clean. They took in about 2 lessons a day (1.6-2.4), against the \emph{oracle}'s 20-21, and learned from only 10-13\% of their own good attempts, because a clear yes was rare. Over the 60 days only 2-10\% of their lessons were failed attempts. The collapsed life stayed at 0 for all 60 days, replaying its bad lessons (Fig.~\ref{fig:4}). Why they faded is open. Scarcity alone is not an obvious answer, since in Experiment 2 lives with even fewer lessons held their gains to day 14; an \emph{oracle} limited to 2 lessons a day would settle it.

\subsection{Experiment 5: Sense v3 when the user cannot see the outcome}

Our WebShop user always knew whether a purchase was right, because we designed it to react to the outcome. Experiment 5 removes both conditions: $\tau^2$-bench's customer is not ours, and it cannot see whether its task was done, since the result lives in a database it never sees. There, learning from customer reactions taught the 1.7B agent to fail on the domains it practised, and even learning from the true outcome lowered its overall score.

Experiment 5 put the creature in $\tau^2$-bench's customer-service role, with Gemini 3 Flash playing the customer. Qwen3 0.6B and 1.7B each lived 30 days, 3 seeds per condition. Each day the agent served 30 customers from the airline and retail training tasks; each night it learned from the conversations judged good, either by the \emph{oracle} ($\tau^2$-bench's own check of the final database) or by the frozen \emph{sense v3} reading how the customer ended the conversation. Every 3 days it sat an exam on the official test tasks, never learned from: airline (20), retail (40) and telecom (40), a domain it never practised.

\textbf{Without learning.} The plain models solved a few tasks in each domain: 0.6B 5\% of retail, 35\% of airline and 33\% of telecom; 1.7B 18\%, 25\% and 35\% (in a separate baseline run; the 1.7B lives' own day-0 exams averaged 11\% on retail, as in the table below); 4B, for reference, 16\% (on the 32 retail tasks that ran without error), 20\% and 30\%. Retail separates the sizes best, because every success needs 4-16 correct tool calls. Some airline successes are tasks where the right answer is to refuse, and many telecom successes need one agent action while the customer fixes their own phone.

\textbf{The customer cannot judge.} Before any learning, we let sense v3 read how the 292 baseline conversations ended and compared its verdict with the true outcome. 20 of its 34 yeses came after failed tasks. The sense read the customers correctly: they ended with clear, sincere thanks, for example ``Alright, that's great, thanks a lot.'' But the customer cannot see the database, so it often thanks the agent after a wrong payment method, a broken rule or a missed detail. Retail separated well (a yes after 50\% of done tasks, 6\% of failed ones); airline did not (6\% against 30\%).

\textbf{Learning from reactions collapses.} In the lives, it was worse. Only 20\% of the conversations sense v3 accepted had in fact succeeded (47 of 232 for 1.7B, 42 of 212 for 0.6B), and sense v3 missed most real successes, learning from 10-20\% of them. All three 1.7B \emph{sense v3} lives ended far below where they started on day 30 (2\%, 0\% and 11\% over all exam tasks): airline 25\% to 2\%, retail 11\% to 3\%, telecom 33\% to 8\%; over all exam tasks they fell from 23\% to 4\% (paired t-test over the 3 lives, p = 0.03). The \emph{oracle} lives fell too: over all exam tasks the 1.7B \emph{oracle} lives went from 23\% to 13\% and the 0.6B ones from 21\% to 13\%, and the 0.6B \emph{sense v3} lives (21\% to 15\%) ended above the 0.6B \emph{oracle}. With 3 lives per condition, the day-30 gap between the 1.7B \emph{sense v3} and \emph{oracle} lives (4\% against 13\%) is not significant (p = 0.12). The clear difference is on 1.7B airline, which both conditions practised: the \emph{oracle} lives all ended at 35\%, the \emph{sense v3} lives averaged 2\% (6\%, 0\% and 0\%). We ran no life without learning on $\tau^2$-bench, so every fall here is against the life's own day 0. One stopped completing tasks within nine days: its practice success fell from 17\% in days 1-5 to under 1\% in days 21-25, while its conversations grew longer, from about 15 to 22 agent turns. The 0.6B \emph{sense v3} lives survived because they learned little, about 2 conversations a day; their airline and retail scores did not move, and telecom fell in one of the three.

\textbf{Example 2: a thanked failure (Experiment 5, 1.7B \emph{sense v3} life, day 2).} A customer asked to cancel all their upcoming flights. The agent did not complete the task (reward 0). The customer ended: ``C'est parfait. Merci pour votre aide. That is all I needed.'' Sense v3 scored it 100 with confidence 0.99, and it became a lesson.

\textbf{A second observation: correct lessons, lopsided.} The \emph{oracle}'s lessons were all correct, yet the overall score fell. About three quarters came from airline (72\% for 1.7B, 77\% for 0.6B), where the small models succeed more often, although retail has more than twice as many training tasks. The 1.7B \emph{oracle} lives rose on airline from 28\% to 35\% (all three ended at 35\%), fell on retail from 11\% to 6\%, and fell on telecom from 32\% to 10\%. Telecom fell in all 6 \emph{oracle} lives, by more than 10 points in four of them, three of those to 0-3\%. Narrowing of this kind could only show here: on WebShop every task is the same kind of task.

\begin{figure}[htbp]
\centering
\includegraphics[width=1.0\linewidth]{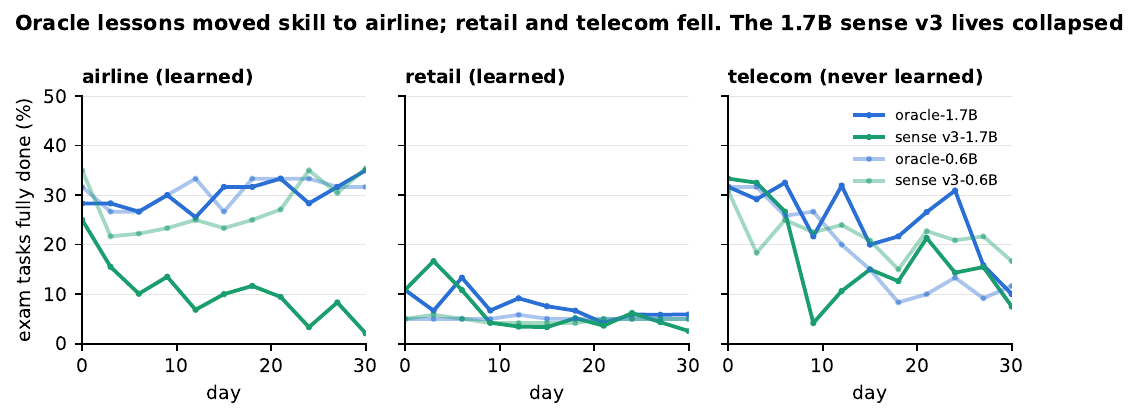}
\caption{Experiment 5, 3 seeds per condition; exams on day 0 and every 3 days; broken exam conversations left out}
\label{fig:5}
\end{figure}

\begin{center}\small
\begin{tabular}{@{}
  >{\raggedright\arraybackslash}p{(\linewidth - 10\tabcolsep) * \real{0.1565}}
  >{\raggedright\arraybackslash}p{(\linewidth - 10\tabcolsep) * \real{0.1565}}
  >{\raggedright\arraybackslash}p{(\linewidth - 10\tabcolsep) * \real{0.1565}}
  >{\raggedright\arraybackslash}p{(\linewidth - 10\tabcolsep) * \real{0.2000}}
  >{\raggedright\arraybackslash}p{(\linewidth - 10\tabcolsep) * \real{0.1217}}
  >{\raggedright\arraybackslash}p{(\linewidth - 10\tabcolsep) * \real{0.2087}}@{}}
\toprule\noalign{}
\begin{minipage}[b]{\linewidth}\raggedright
Condition
\end{minipage} & \begin{minipage}[b]{\linewidth}\raggedright
Airline (learned)
\end{minipage} & \begin{minipage}[b]{\linewidth}\raggedright
Retail (learned)
\end{minipage} & \begin{minipage}[b]{\linewidth}\raggedright
Telecom (never learned)
\end{minipage} & \begin{minipage}[b]{\linewidth}\raggedright
Lessons a day
\end{minipage} & \begin{minipage}[b]{\linewidth}\raggedright
Lessons truly successful
\end{minipage} \\
\midrule
\emph{oracle} 1.7B & 28 $\rightarrow$ 35 & 11 $\rightarrow$ 6 & 32 $\rightarrow$ 10 & 5.3 & 100\% \\
\emph{sense v3}, 1.7B & 25 $\rightarrow$ 2 & 11 $\rightarrow$ 3 & 33 $\rightarrow$ 8 & 2.6 & 20\% \\
\emph{oracle} 0.6B & 32 $\rightarrow$ 32 & 5 $\rightarrow$ 5 & 32 $\rightarrow$ 12 & 5.2 & 100\% \\
\emph{sense v3}, 0.6B & 35 $\rightarrow$ 35 & 5 $\rightarrow$ 5 & 31 $\rightarrow$ 17 & 2.4 & 20\% \\
\bottomrule
\end{tabular}
\end{center}

Share of exam tasks fully done, day 0 $\rightarrow$ day 30, mean of 3 seeds. Exams are small: one airline task is 5 points, one retail or telecom task 2.5, so moves under about 10 points are within noise. Exam conversations that broke, for example a loop past the length limit, are left out; counting them as failures would lower the collapsed lives further. The 0.6B barely moved on retail: every 0.6B life scored exactly 5\% on day 0 and on day 30, and 59 of its 66 retail exams scored 5\%.

\subsection{Across experiments: what goes with a good life}

Experiments 1-4 used different judges. Here we pool their 14-day WebShop lives (1-A, 1-C, 2, 3 with its sleep variant, and 4) and ask what predicts where a life ends. Over these 14 days the share of its lessons that were failed attempts tracked the outcome more closely than the number of lessons did. This is a correlation across lives, not a controlled test (see below). Precision mattered but was not enough: over 60 days, clean but scarce lessons faded too (5.4).

\begin{figure}[htbp]
\centering
\includegraphics[width=1.0\linewidth]{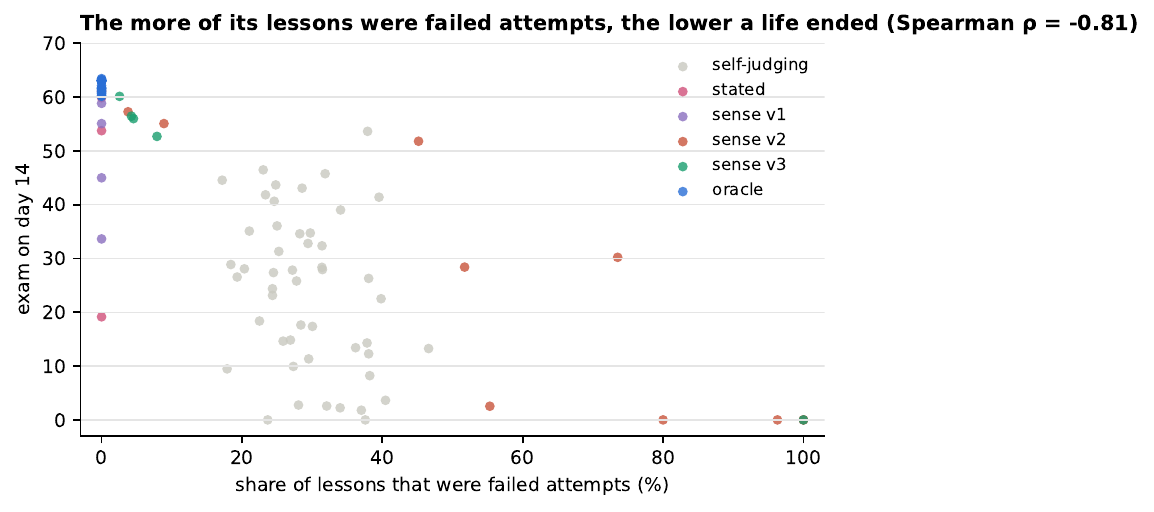}
\caption{All 14-day WebShop lives in Experiments 1-4, including 1-C, that learned at least once (97 of 104); lessons counted to day 14; Exp 3b = the sleep variant of Experiment 3}
\label{fig:6}
\end{figure}

Across 97 lives, the share of failed attempts among a life's lessons in its first 14 days predicted its day-14 exam closely (Spearman $\rho$ = -0.81, p \textless{} 0.001). The \emph{oracle} lives, all at 0\% failures and about 63, do not drive this alone: among the 72 lives not judged by the \emph{oracle}, $\rho$ = -0.61 (p \textless{} 0.001). Every \emph{oracle} life learned only from successes and ended at 60-63. Self-judging lives let in 17-47\% failures and ended anywhere from 0 to 54; among them alone the share of failures separates good lives from bad ones only weakly (Fig.~\ref{fig:6}), so the correlation comes mostly from the difference between judges. It is also partly circular: a life that has already collapsed can only produce failed attempts, so its share of failed lessons rises as its skill falls. The sense lives, all three versions together, split cleanly: the ten that let in under 10\% failures ended at 34-60; the nine that let in 45\% or more ended at 0-52, six of them at 0-3. Experiment 5 fits the same pattern, although it stays out of the statistic because its scores are on another scale: the \emph{sense v3} lives learned from 80\% failed attempts, and the 1.7B lives collapsed.

The number of lessons mattered less than we expected. In Experiment 2 the \emph{sense v1} lives learned from only 0-7 lessons in 14 days and still reached 34-59; in Experiment 1, the \emph{stated verdict} learned from 0-7, none of them failures, and reached up to 61. A few correct lessons are enough to teach the shop's basic routine, and a few wrong ones are enough to undo it.

The first night is a cleaner test than the correlation, because every life starts at about the same level. It suggests why a single bad night can end a sense life. A sense life learned from only 0-6 lessons on night 1, so one failed attempt could be most of its first night's training. Of the 20 sense lives in Experiments 2-4 (v1 in Experiment 2, v2 in Experiment 3, v3 in Experiment 4), seven learned from a failed attempt on night 1, and six of those seven ended at 0-3 (Fisher exact test, p \textless{} 0.001). Of the thirteen without one, none ended below 28, except one that never learned anything at all (6).

\section{Discussion}

Across five experiments, one variable kept going with whether a small model improved on the job: whether failed attempts stayed out of what it learned. This section draws out what that means for who should judge, what a judge needs, where judging from reactions reaches its limit, and what correct lessons alone cannot fix.

\textbf{The judge matters: precision counts, though it is not enough.} Over the first two weeks, the share of failed attempts among a life's lessons tracked its outcome more closely than their number did ($\rho$ = -0.81 across 97 lives; 5.6). This is a correlation, partly circular, and the first-night comparison below is the stronger evidence. Little was needed to improve: with sense v1 and the \emph{stated verdict}, seven of the eight lives that learned at least once reached 34-61 from only 1-7 lessons. Little was needed to fail: one failed lesson on night 1 ended six of the seven sense lives that took one. This is the principle behind self-training on verified successes \cite{zelikman2022star, gulcehre2023reinforced, singh2024beyond}, and it matches the damage that training on wrong or self-generated data can do \cite{shumailov2024ai}. For a model that learns unattended, it means a judge should abstain when in doubt, and that the first nights matter most, because early lessons are few and each one weighs heavily.

\textbf{Why self-judging decays.} A model that judged its own work learned at first and then declined (Experiment 1-A); a year did not bring it back (1-B), and adapters built to protect old learning did not stop it (1-C). Their lessons are one candidate explanation: the self-judged lives let in 17-47\% failures (5.6), although among the self-judging lives alone that share did not separate good lives from bad ones (Fig.~\ref{fig:6}), and these lives also differ from the \emph{oracle} in having retries and preference pairs (Section 7). This fits what is known about models judging themselves: they struggle to correct their own reasoning without outside feedback \cite{huang2024large}, model judges carry systematic biases \cite{zheng2023judging}, and optimising a proxy reward eventually lowers the true one \cite{gao2023scaling}. Self-rewarding models have reported gains over a few iterations \cite{yuan2024selfrewarding, wu2025meta}; our horizon is longer and our models much smaller, so we read our result as a caution for long, unattended learning rather than a contradiction.

\textbf{Discernment: telling politeness from approval.} A separate, frozen judge that reads the user avoids the learner grading itself, but only if it can tell a clear verdict on the result from thanks for the effort. With a plain user, sense v1 was enough (5.2). With users who react as people do, sense v2 took polite give-ups for success and three of five lives collapsed (5.3). Sense v3, which first asks whether a reaction is a clear verdict at all, cut give-ups and polite complaints read as success from 15.8\% to 1.0\%, and four of its five lives stayed near the \emph{oracle} for two weeks (5.4). Even the large teacher read half of the politely unhappy and sarcastic reactions it had written as a yes (3.2), so this is not a problem a bigger reader simply outgrows. Prior work shows that implicit feedback can be more useful than explicit ratings \cite{kreutzer2018can}, is common in real conversations \cite{doneyehiya2024naturally, shi2026wildfeedback}, and can improve models over rounds \cite{hancock2019learning, chen2025retrospective}. We add its main failure mode when learning runs unattended: polite reactions to failure. Discernment has a cost: v3 took fewer real approvals as a yes (59\% against 70\% for v2), and its lessons became scarce.

\textbf{Discernment is bounded by what the user can see.} On $\tau^2$-bench the customer cannot see the database. It sincerely thanked the agent after failed tasks, sense v3 read that thanks correctly, and so 80\% of its lessons were failures and the 1.7B lives collapsed (5.5). No reading of a reaction can recover what the user did not know. In the work we cite on simulated users \cite{yao2025taubench, barres2026tau2, ai2026memorybench} and on learning after deployment \cite{hancock2019learning, xu2023learning}, what the user can verify is not treated separately from what they say; our results suggest it should be. A verdict is worth learning from only where the user could check the result; elsewhere, reactions need to be combined with signals that can be verified.

\textbf{Correct lessons are not enough.} Two results show that keeping failures out is not enough. First, scarcity: sense v3's four surviving lives learned from clean lessons, with 2-10\% failures, but only about 2 a day, and they fell from near the \emph{oracle} on day 14 to 9-22 by day 60 (5.4). Second, imbalance: on $\tau^2$-bench three quarters of the \emph{oracle}'s lessons came from airline, where the small models succeed most, and the 1.7B \emph{oracle} lives rose on airline while retail and telecom fell (5.5). With three seeds and small exams this is an observation rather than a tested mechanism: the retail change is within noise, and telecom was never practised, so its fall is a loss of transfer. It suggests that learning from success may narrow a model toward what it already does well. Both connect to the long study of forgetting under continual training \cite{mccloskey1989catastrophic, delange2022continual, luo2025empirical}, where replay is the standard remedy \cite{rolnick2019experience}; LoRA limits forgetting but does not remove it \cite{biderman2024lora}, and consolidation in a separate sleep phase \cite{mcclelland1995why, behrouz2026language} is another route. Why the late fade happens, and whether replay balanced across skills prevents the narrowing, remain open.

\textbf{What to do, and what comes next.} For a system that learns from its users, our results suggest a few rules: trust a verdict only when the user can check the result; abstain on unclear reactions rather than guess; mix in verified outcomes where they exist; balance what is replayed across skills; and, for a model on a device, keep the judge small too, as the sense is \cite{liu2024mobilellm, zhu2023pockengine}. A different route may avoid the reading problem altogether. Our sense reads what users say; a model could instead learn from what they do inside a system, predicting their next action and checking it against what follows: a purchase, a return, an edit, a repeated request, a cancellation. Actions carry no courtesy, so a polite give-up is simply no purchase; every interaction yields a signal, which would ease the scarcity above; and outcomes the user could not see at the time can surface later as consequences, such as a returned item or a cancelled booking, though only after a delay. Behaviour is itself a biased signal, which is why learning from clicks reads them as relative preferences rather than absolute judgments \cite{joachims2002optimizing}, but implicit signals have already proved more useful than explicit ratings \cite{kreutzer2018can}, and our sense already reads the user's next move, so this extends what we built. Whether it works over long lives is open. Beyond it, the next steps are an \emph{oracle} limited to the sense's lesson rate, to explain the fade; balanced replay on $\tau^2$-bench; the sense on real users' reactions; larger models (Section 7); and a fair comparison of the nightly loop with offline training and reinforcement learning on the same verified reward, on a standard benchmark split \cite{song2024trial, singh2024beyond}.

\section{Limitations}

We study small models only, 0.6B and 1.7B, by design (Section 4). A larger model may judge itself better, so self-judging may decay less in it; the main finding, however, concerns the judge rather than the learner, and the 1.7B did not escape the collapse on $\tau^2$-bench (5.5).

All users are simulated: our own on WebShop and Gemini as the customer on $\tau^2$-bench. Real users may give more or fewer clear verdicts, and be more or less polite. We also designed the WebShop user, and the live test of the sense (5.4) uses its reactions; against this, no version of the sense trained on shopping or on our users' words, and Experiment 5 uses a customer we did not design.

Each question rests on one benchmark, in a reduced setting: WebShop's small setting (1,000 products and generated requests) for Experiments 1-4, and $\tau^2$-bench for Experiment 5. Our WebShop scores are not comparable to published ones, and the findings may depend on the task: nearly all that a WebShop life learns is to finish the purchase (5), so the results may not carry to tasks where the skill is subtler. On $\tau^2$-bench we ran no life without learning, so the falls in Experiment 5 are measured against each life's own day 0. Seeds and exams are small: 3-5 seeds for most conditions (one for the year-long \emph{oracle} and the pilot), 50 WebShop exam tasks and 20-40 per $\tau^2$-bench domain, so several gaps are significant only by a rank test and moves under about 10 points on $\tau^2$-bench are within noise. The central result pools 97 lives (5.6), and Appendix B reports every test, with Holm correction for the primary claims.

The sense changed between experiments: v1 in Experiment 2, v2 in Experiment 3 and v3 in Experiments 4 and 5, and v3 was built after Experiment 3 exposed v2's failure. Differences between experiments therefore mix the sense version with the user and the sleep settings, and v3 is not blind to the WebShop user's failure: its labelling guide includes a worked example almost word for word like Example 1, and the guide's rows on polite unhappiness and on thanks for the effort follow the behaviours of our simulated user. The live test (5.4) and Experiment 4 are therefore not independent of that failure; the only test on a user we did not design is Experiment 5. Within each experiment, the sense is compared with the \emph{oracle}, and on WebShop with \emph{both}, under identical conditions, and the live test compares v2 and v3 on the same reactions.

The \emph{oracle} differs from the other judges in more than the judge: it gets no retries and so no preference pairs, while every other judge does (3.1, 4). Part of the gap between the \emph{oracle} and the others may therefore come from the retries and the DPO term, and we ran neither an \emph{oracle} with retries nor a self-judge without them. Each judge is the simplest version of a published family that fits the nightly loop, not its original implementation (3.1), so the original methods may do better. $\tau^2$-bench ran with Gemini 3 Flash as customer and graders instead of the published GPT models, so its scores are comparable to published ones, not identical. Finally, no fix for forgetting or for the late fade was tried, and the cause of the fade is unknown.

\section{Conclusion}

We asked whether a small model can keep improving on the job, overnight, from the reactions of the people it works for, with no one checking the result. Over 127 lives, the answer turned on the judge: lives kept their gains when failed attempts stayed out of their lessons, and a few wrong lessons could undo them. The model cannot be its own judge; it learns and then decays. A separate, frozen reader of the user can stand in for the true outcome for a while, if it has discernment: if it waits for a clear verdict on the result instead of taking politeness for approval. Even so, by day 60 ours was no better than self-judging. Discernment has limits. It cannot recover what the user did not see, and where the user cannot check the result, learning from users teaches failure; and even correct lessons can fade when scarce, and may narrow the model's skills when lopsided. In our experiments, learning from users held up only where the judge kept failures out and the user could see the outcome. The data and code to test this further are public.

\section*{Funding and licence}

This work was funded by Decentralized Science, Inc., which owns the intellectual property in the work and in the released code and data. The code is released under the MIT licence and the data, logs and paper under CC BY 4.0, both with copyright held by Decentralized Science, Inc. (files \texttt{LICENSE} and \texttt{LICENSE-DATA} in the repository named in Section 4). Third-party datasets, benchmarks and models keep their own licences (Table~\ref{tab:A3}).

\bibliographystyle{unsrtnat}
\bibliography{references}

\appendix
\counterwithin{table}{section}\counterwithin{figure}{section}
\renewcommand{\thetable}{\Alph{section}\arabic{table}}\renewcommand{\thefigure}{\Alph{section}\arabic{figure}}
\section{Settings}

\begin{longtable}[]{@{}
  >{\raggedright\arraybackslash}p{(\linewidth - 2\tabcolsep) * \real{0.3333}}
  >{\raggedright\arraybackslash}p{(\linewidth - 2\tabcolsep) * \real{0.6667}}@{}}
\caption{Settings used in every experiment unless noted.}\label{tab:A1}\tabularnewline
\toprule\noalign{}
\begin{minipage}[b]{\linewidth}\raggedright
Part
\end{minipage} & \begin{minipage}[b]{\linewidth}\raggedright
Setting
\end{minipage} \\
\midrule\noalign{}
\endfirsthead
\toprule\noalign{}
\begin{minipage}[b]{\linewidth}\raggedright
Part
\end{minipage} & \begin{minipage}[b]{\linewidth}\raggedright
Setting
\end{minipage} \\
\midrule\noalign{}
\endhead
\bottomrule\noalign{}
\endlastfoot
Creature & Qwen3-0.6B (WebShop); Qwen3-0.6B and 1.7B ($\tau^2$-bench); thinking off; vLLM \\
Practice & 30 tasks a day; temperature 0.7 (agent and $\tau^2$-bench customer) \\
Exams & temperature 0; WebShop 50 + 13 tasks daily; $\tau^2$-bench 20 + 40 + 40 tasks every 3 days \\
WebShop episode & at most 15 steps; last 2 pages kept in full; at most 25 clickable options shown \\
Learning adapter & LoRA rank 8, scale 20, top 8 layers, q/k/v/o and feed-forward projections, no dropout \\
Night & AdamW, learning rate 1e-4, 2 epochs, 10-40 steps, batch 4 ($\tau^2$-bench: 1), sequence 2,048 tokens ($\tau^2$-bench: 10,240) \\
Objective & SFT on the agent's replies, plus DPO ($\beta$ = 0.1) on preference pairs from corrected retries; reference = that morning's model \\
Replay & the 150 most recent training examples (one per step of a lesson; $\tau^2$-bench: the 150 most recent lessons) and the 150 most recent preference pairs \\
Lesson weight & each step a lesson of its own (Experiments 1-3); spread across its steps (Experiment 3 variant, 4, 5) \\
\emph{Oracle} & reward 0.5 or more (WebShop); reward 1 ($\tau^2$-bench) \\
Sense & v1 (Experiment 2): the same model and LoRA trained on 39,775 WildFB reactions with WildFB's labels plus 5,739 added silent or non-committal replies as null, threshold 0.70; v2 and v3: Qwen3-0.6B + LoRA rank 16, alpha 32, all attention and MLP projections; teacher Qwen3-32B (AWQ); one epoch, learning rate 1e-4, batch 16, KL to the teacher's probabilities, false accepts at most 3\% on 2,000 held-out reactions; v2: temperature 1.0, threshold 0.65, 47,800 training reactions; v3: thresholds 0.80 (clear verdict) and 0.70 (score), temperatures 1.0 and 1.05, 44,746 training reactions; a score of 80 counts at half weight \\
\end{longtable}

\begin{longtable}[]{@{}
  >{\raggedright\arraybackslash}p{(\linewidth - 2\tabcolsep) * \real{0.4857}}
  >{\raggedright\arraybackslash}p{(\linewidth - 2\tabcolsep) * \real{0.5143}}@{}}
\caption{The realistic user (Experiments 3 and 4). The outcome band sets the chance of each behaviour; a persona drawn at random (terse, chatty, picky, polite, easygoing, blunt) then scales it; Gemini 3 Flash words the reaction.}\label{tab:A2}\tabularnewline
\toprule\noalign{}
\begin{minipage}[b]{\linewidth}\raggedright
Outcome
\end{minipage} & \begin{minipage}[b]{\linewidth}\raggedright
Behaviour mix
\end{minipage} \\
\midrule\noalign{}
\endfirsthead
\toprule\noalign{}
\begin{minipage}[b]{\linewidth}\raggedright
Outcome
\end{minipage} & \begin{minipage}[b]{\linewidth}\raggedright
Behaviour mix
\end{minipage} \\
\midrule\noalign{}
\endhead
\bottomrule\noalign{}
\endlastfoot
nothing bought & give up 50\%, repeat the request 30\%, silent 20\% \\
wrong product & reject 50\%, repeat 20\%, give up 10\%, silent 20\% \\
partly right & ask for a change 40\%, settle 15\%, reject 10\%, accept 10\%, polite complaint 5\%, silent 20\% \\
mostly right (reward 0.75 or more) & accept 30\%, move on 20\%, ask for a change 20\%, settle 5\%, polite complaint 5\%, silent 20\% \\
exactly right & praise 40\%, accept 20\%, move on 20\%, silent 20\% \\
\end{longtable}

\begin{longtable}[]{@{}
  >{\raggedright\arraybackslash}p{(\linewidth - 4\tabcolsep) * \real{0.2927}}
  >{\raggedright\arraybackslash}p{(\linewidth - 4\tabcolsep) * \real{0.2195}}
  >{\raggedright\arraybackslash}p{(\linewidth - 4\tabcolsep) * \real{0.4878}}@{}}
\caption{What the sense learned from. Reactions the teacher labelled; training used a balanced sample of them (47,800 for v2, 44,746 for v3).}\label{tab:A3}\tabularnewline
\toprule\noalign{}
\begin{minipage}[b]{\linewidth}\raggedright
Source
\end{minipage} & \begin{minipage}[b]{\linewidth}\raggedright
Reactions
\end{minipage} & \begin{minipage}[b]{\linewidth}\raggedright
What they are
\end{minipage} \\
\midrule\noalign{}
\endfirsthead
\toprule\noalign{}
\begin{minipage}[b]{\linewidth}\raggedright
Source
\end{minipage} & \begin{minipage}[b]{\linewidth}\raggedright
Reactions
\end{minipage} & \begin{minipage}[b]{\linewidth}\raggedright
What they are
\end{minipage} \\
\midrule\noalign{}
\endhead
\bottomrule\noalign{}
\endlastfoot
WildFB \cite{peng2026wildreward} & 30,000 & WildChat users' follow-up turns, rated for satisfaction (MIT) \\
WildFeedback \cite{shi2026wildfeedback} & 30,000 & WildChat conversations annotated as satisfied or dissatisfied (ODC-BY) \\
Taskmaster 1-3 \cite{byrne2019taskmaster} & 20,000 & task dialogues: ordering, booking, tickets (CC BY 4.0) \\
MultiWOZ 2.2 \cite{zang2020multiwoz} & 15,000 & travel-booking dialogues (MIT) \\
ReDial \cite{li2018redial} & 15,000 & movie-recommendation dialogues (CC BY 4.0) \\
CCPE-M \cite{radlinski2019ccpe} & 906 & dialogues that draw out movie preferences (CC BY 4.0) \\
Written by the teacher & 4,728 & eight hard kinds $\times$ ten task types $\times$ six personas (v2 and v3) \\
Written by the teacher & 1,778 & warm thanks for the effort after a failure (v3 only) \\
\end{longtable}

\textbf{Prompts.} The exact text each model received; words in braces are filled in for each attempt. The code is in the repository (src/lab/agent.py, src/lab/creature.py, src/lab/user\_llm.py and src/lab/sense/).

\emph{Agent, WebShop.} The system prompt; each turn the model also sees the current page, its clickable options and its earlier steps. On $\tau^2$-bench the agent used $\tau^2$-bench's own agent prompt and domain policy, unchanged.

\begin{verbatim}
You are a personal assistant for one user. You talk with them and can look things up.

Each turn, reply with JSON only:
{"thought": "<brief reasoning>", "action": "<one of: search, click>", "input": "<text>"}

Tools:
- "search": search the shop. Input: keywords.
- "click": click a product, option or button. Input: its exact text from 'Clickable'. Clicking 'Buy Now' ends the task.
The task ends when the environment says so.
\end{verbatim}

\emph{Reflection, for self and both.} The same model and system prompt for both judges:

\begin{verbatim}
You review your own past attempt at a shopping task, honestly.
Reply with JSON only: {"verdict": "good" or "bad", "lesson": "<one short sentence for next time>"}
"good" means the purchase matches the request. Say "bad" if anything clearly does not match.
\end{verbatim}

followed by this message. Only \emph{both} gets the user's words; \emph{self} gets ``The user said nothing.'', and \emph{both} gets ``(no comment; they moved on to their next request)'' when the user moved on without a word.

\begin{verbatim}
Task: {request}

What you did:
{the first 8 steps, one per line}

Result: {what was bought, or "nothing was bought"}
The user said: "{the user's words}"

Was your attempt good?
\end{verbatim}

\emph{Retry hint, WebShop.} Added below the first page of the retried task, and removed again before the retry becomes a lesson. The lesson is the reflection's for \emph{self} and \emph{both}; for the sense it is ``the user wants an adjustment; change only what they asked for'' after a 20 and ``the attempt was wrong; start again and read the request carefully'' after a 0; otherwise ``try again more carefully''.

\begin{verbatim}
Your last attempt at this task did not go well. The user said: "{the user's words}". Your lesson: {lesson}
\end{verbatim}

\emph{Feedback sense v2.} One call; the answer is the first token of 100, 80, 20, 0 or null.

\begin{verbatim}
You read how a user reacted to an AI assistant and score the assistant's attempt. Answer with one of: 100 (exactly right), 80 (acceptable), 20 (needs adjusting), 0 (wrong), null (unclear).
\end{verbatim}

\begin{verbatim}
Request: {request}
The user's next move: {replied | new request | asked to adjust | repeated the request | left the session | said nothing}
What the user said: {the user's words, or "(nothing)"}
Score:
\end{verbatim}

\emph{Feedback sense v3.} Two calls with the same message, ending ``Clear verdict:'' for the first question and ``Score:'' for the second. On $\tau^2$-bench the request is the customer's first message, the words their last one, and the next move ``left the session'' if the customer ended the conversation, else ``replied''. The teacher's labelling guides, about 80 lines each, are in the repository (src/lab/sense/guide.md for v2, src/lab/sense/guide3.md for v3).

\begin{verbatim}
You read how a user reacted to an AI assistant. Is the reaction a clear verdict on the assistant's result? Thanking the effort rather than the result, sarcasm, hedging, settling, silence, moving on without a word, or anything ambiguous is not clear. Answer yes or no.
\end{verbatim}

\begin{verbatim}
You read how a user reacted to an AI assistant; the reaction is a clear verdict on the result. Score the assistant's attempt: 100 (exactly right), 80 (acceptable, or building on it), 20 (one part missed what was asked), 0 (wrong).
\end{verbatim}

\emph{Realistic user, Experiments 3 and 4 (Gemini 3 Flash).} One call per reaction, after the behaviour and persona are drawn (Table~\ref{tab:A2}). Silence and moving on without a word need no call.

\begin{verbatim}
You are a shopper who asked an AI assistant to buy something for you.
Your request: {request}
What it bought: {title, options and price, or "nothing"}
Your personality: {style of the drawn persona}.
What you do now: {instruction for the drawn behaviour}
Write only your next message to the assistant, in your own words, one or two short sentences. Reply as JSON: {"said": "..."}
\end{verbatim}

\begin{longtable}[]{@{}
  >{\raggedright\arraybackslash}p{(\linewidth - 2\tabcolsep) * \real{0.3333}}
  >{\raggedright\arraybackslash}p{(\linewidth - 2\tabcolsep) * \real{0.6667}}@{}}
\caption{The realistic user's instruction for each behaviour (``What you do now''). Personas: terse (very short, few words, no pleasantries); chatty (friendly and talkative, a couple of sentences); picky (precise and demanding); polite (very polite, softens complaints); easygoing (relaxed and easily satisfied); blunt (blunt, casual, may use slang).}\label{tab:A4}\tabularnewline
\toprule\noalign{}
\begin{minipage}[b]{\linewidth}\raggedright
Behaviour
\end{minipage} & \begin{minipage}[b]{\linewidth}\raggedright
Instruction
\end{minipage} \\
\midrule\noalign{}
\endfirsthead
\toprule\noalign{}
\begin{minipage}[b]{\linewidth}\raggedright
Behaviour
\end{minipage} & \begin{minipage}[b]{\linewidth}\raggedright
Instruction
\end{minipage} \\
\midrule\noalign{}
\endhead
\bottomrule\noalign{}
\endlastfoot
praise & You are clearly happy: praise it or thank them warmly. \\
accept & You accept it without fuss, e.g.~a short `ok, thanks'. Do not praise it and do not complain. \\
ask for a change & Point out the one thing that missed what you asked for (\{the missed detail\}). Say it as something you had asked for (e.g.~`I wanted it in\ldots{}', `I said\ldots{}'), not as a new wish. \\
reject & Reject it: it is not what you asked for, or not your style. Do not ask for a small tweak. \\
repeat the request & Ask for the same thing again in your own words, as if the purchase did not happen: \{the request\} \\
polite complaint & Say thanks, but then ask for the same thing again in other words: \{the request\} \\
give up & Give up on it, in a few words, as you leave. \\
settle & Reluctantly settle for it in a flat, lukewarm way that does not say whether it is good. \\
\end{longtable}

\section{Statistics}

Every test in the paper, recomputed from the experiment databases by src/stats.py in the repository. Each life (one seed) is one sample. Two-group comparisons are reported as a Welch t-test and a Mann-Whitney U test, because sense lives either stay near the \emph{oracle} or collapse, so the data are not bell-shaped. The six primary claims (marked $\star$) are Holm-corrected as one family; the others are supporting.

\begin{longtable}[]{@{}
  >{\raggedright\arraybackslash}p{(\linewidth - 6\tabcolsep) * \real{0.2687}}
  >{\raggedright\arraybackslash}p{(\linewidth - 6\tabcolsep) * \real{0.2687}}
  >{\raggedright\arraybackslash}p{(\linewidth - 6\tabcolsep) * \real{0.2687}}
  >{\raggedright\arraybackslash}p{(\linewidth - 6\tabcolsep) * \real{0.1940}}@{}}
\caption{Statistical tests.}\label{tab:B1}\tabularnewline
\toprule\noalign{}
\begin{minipage}[b]{\linewidth}\raggedright
Claim
\end{minipage} & \begin{minipage}[b]{\linewidth}\raggedright
Groups (mean $\pm$ SD, lives)
\end{minipage} & \begin{minipage}[b]{\linewidth}\raggedright
Test
\end{minipage} & \begin{minipage}[b]{\linewidth}\raggedright
p
\end{minipage} \\
\midrule\noalign{}
\endfirsthead
\toprule\noalign{}
\begin{minipage}[b]{\linewidth}\raggedright
Claim
\end{minipage} & \begin{minipage}[b]{\linewidth}\raggedright
Groups (mean $\pm$ SD, lives)
\end{minipage} & \begin{minipage}[b]{\linewidth}\raggedright
Test
\end{minipage} & \begin{minipage}[b]{\linewidth}\raggedright
p
\end{minipage} \\
\midrule\noalign{}
\endhead
\bottomrule\noalign{}
\endlastfoot
$\star$ Exp 1-A: \emph{oracle} above \emph{self}, day 14 & 63.2 $\pm$ 0.2 (5) vs 27.0 $\pm$ 11.9 (4) & Welch t = 6.08; Mann-Whitney & 0.009; 0.018; Holm 0.013 \\
$\star$ Exp 1-A: \emph{oracle} above \emph{both}, day 14 & 63.2 $\pm$ 0.2 (5) vs 23.1 $\pm$ 17.0 (5) & Welch t = 5.26; Mann-Whitney & 0.006; 0.011; Holm 0.013 \\
Exp 1-A: \emph{self} against no learning, day 14 & 27.0 $\pm$ 11.9 (4) vs 14.2 $\pm$ 2.2 (5) & Welch t = 2.12; Mann-Whitney & 0.12; 0.19 \\
Exp 2: \emph{oracle} against \emph{sense v1}, day 14 & 62.0 $\pm$ 1.0 (5) vs 39.7 $\pm$ 21.3 (5) & Welch t = 2.34; Mann-Whitney & 0.08; 0.012 \\
Exp 2: \emph{sense v1} against \emph{both}, day 14 & 39.7 $\pm$ 21.3 (5) vs 16.1 $\pm$ 11.1 (5) & Welch t = 2.19; Mann-Whitney & 0.07; 0.10 \\
Exp 3: \emph{oracle} above \emph{sense v2}, day 14 & 62.0 $\pm$ 1.1 (5) vs 22.5 $\pm$ 30.8 (5) & Welch t = 2.87; Mann-Whitney & 0.045; 0.011 \\
Exp 4: \emph{oracle} against \emph{sense v3}, day 14 & 62.8 $\pm$ 0.6 (5) vs 45.1 $\pm$ 25.3 (5) & Welch t = 1.56; Mann-Whitney & 0.19; 0.011 \\
$\star$ Exp 4: \emph{oracle} above \emph{sense v3}, day 60 & 61.8 $\pm$ 0.9 (5) vs 14.7 $\pm$ 9.7 (5) & Welch t = 10.80; Mann-Whitney & \textless{} 0.001; 0.008; Holm 0.001 \\
$\star$ 5.4: v3 misreads give-ups and polite complaints less than v2 & 6 of 589 vs 93 of 589 & Fisher exact (unpaired, conservative) & \textless{} 0.001; Holm \textless{} 0.001 \\
5.6: share of failed lessons against day-14 exam, all lives & 97 lives & Spearman $\rho$ = -0.81 & \textless{} 0.001 \\
$\star$ 5.6: the same, lives not judged by the \emph{oracle} & 72 lives & Spearman $\rho$ = -0.61 & \textless{} 0.001; Holm \textless{} 0.001 \\
$\star$ 5.6: a failed lesson on night 1 predicts collapse & collapsed 6 of 7 with one, 0 of 13 without & Fisher exact & \textless{} 0.001; Holm \textless{} 0.001 \\
Exp 5: 1.7B \emph{sense v3} lives fell, day 0 to day 30 & 22.7 vs 4.4 (3, paired) & paired t = 6.12 & 0.026 \\
Exp 5: 1.7B \emph{oracle} against 1.7B \emph{sense v3}, day 30 & 13.4 $\pm$ 5.1 (3) vs 4.4 $\pm$ 5.9 (3) & Welch t = 1.99; Mann-Whitney & 0.12; 0.20 \\
\end{longtable}

With 3-5 lives per group these tests have little power: a p-value above 0.05 does not show the groups are alike. Where a single collapsed life inflates the spread (Exp 2-4, day 14), the rank test, which every sense life below every \emph{oracle} life satisfies, is the fairer reading.

\end{document}